\documentclass[journal]{IEEEtran}
\usepackage{cite}

\usepackage{graphicx}
\graphicspath{ {images/} }

\usepackage[cmex10]{amsmath}
\usepackage{algorithmicx}

\usepackage{array}

\ifCLASSOPTIONcompsoc
  \usepackage[caption=false,font=normalsize,labelfont=sf,textfont=sf]{subfig}
\else
  \usepackage[caption=false,font=footnotesize]{subfig}
\fi
\usepackage{url}

\usepackage{amsfonts}
\usepackage{epstopdf}
\usepackage{algorithm}

\newcommand{\bD}{\mathbf{D}}
\newcommand{\bd}{\mathbf{d}}
\newcommand{\br}{\mathbf{r}}
\newcommand{\bx}{\mathbf{x}}

\newcommand{\bk}{\mathbf{k}}

\newcommand{\bq}{\mathbf{q}}
\newcommand{\bbf}{\mathbf{f}}
\newcommand{\bz}{\mathbf{z}}

\newcommand{\bh}{\mathbf{h}}
\newcommand{\ba}{\mathbf{a}}

\newcommand{\bv}{\mathbf{v}}
\newcommand{\bX}{\mathbf{X}}
\newcommand{\cX}{\mathcal{X}}

\newcommand{\cF}{\mathcal{F}}

\newcommand{\by}{\mathbf{y}}

\newcommand{\bK}{\mathbf{K}}
\newcommand{\bF}{\mathbf{F}}

\newcommand{\bT}{\mathbf{T}}
\newcommand{\bC}{\mathbf{C}}
\newcommand{\bH}{\mathbf{H}}

\newcommand{\bW}{\mathbf{W}}
\newcommand{\bQ}{\mathbf{Q}}

\newcommand{\bS}{\mathbf{S}}
\newcommand{\bB}{\mathbf{B}}
\newcommand{\bA}{\mathbf{A}}
\newcommand{\bI}{\mathbf{I}}
\newcommand{\bU}{\mathbf{U}}
\newcommand{\bV}{\mathbf{V}}
\newcommand{\bGamma}{\boldsymbol{\Gamma}}
\newcommand{\bTheta}{\boldsymbol{\Theta}}
\newcommand{\bgamma}{\boldsymbol{\gamma}}

\newcommand{\bmu}{\boldsymbol{\mu}}

\begin{document}
%
\title{Linearized Kernel Dictionary Learning}
%
%
%

\author{Alona~Golts~and~Michael~Elad,~\IEEEmembership{IEEE~Fellow}
       \thanks{The research leading to these results has received funding from the
European Research Council under European Unions Seventh Framework
Program, ERC Grant agreement no. 320649.}
}

%
%

\markboth{}%
{Golts \MakeLowercase{\textit{et al.}}: Linearized Kernel Dictionary Learning}
%



\maketitle

\begin{abstract}
In this paper we present a new approach of incorporating kernels into dictionary learning. The kernel K-SVD algorithm (KKSVD), which has been introduced recently, shows an improvement in classification performance, with relation to its linear counterpart K-SVD. However, this algorithm requires the storage and handling of a very large kernel matrix, which leads to high computational cost, while also limiting its use to setups with small number of training examples. We address these problems by combining two ideas: first we approximate the kernel matrix using a cleverly sampled subset of its columns using the Nystr\"{o}m method; secondly, as we wish to avoid using this matrix altogether, we decompose it by SVD to form new ``virtual samples'', on which any linear dictionary learning can be employed. Our method, termed ``Linearized Kernel Dictionary Learning'' (LKDL) can be seamlessly applied as a pre-processing stage on top of any efficient off-the-shelf dictionary learning scheme, effectively ``kernelizing'' it. We demonstrate the effectiveness of our method on several tasks of both supervised and unsupervised classification and show the efficiency of the proposed scheme, its easy integration and performance boosting properties.
\end{abstract}

\begin{IEEEkeywords}
Dictionary Learning, Supervised Dictionary Learning, Kernel Dictionary Learning, Kernels, KSVD.
\end{IEEEkeywords}

%
\IEEEpeerreviewmaketitle

\section{Introduction}

\IEEEPARstart{T}{he} field of sparse representations has witnessed great success in an array of applications in signal and image processing.
The basic operation in sparse representations is called ``sparse coding'', which involves the reconstruction of the signals of interest using a sparse set of building blocks, referred to as ``atoms''. The atoms are gathered in a structure called the ``dictionary'', which can be manually crafted to contain mathematical functions that are proven successful in representing signals and images, such as wavelets \cite{WaveletDic}, curvelets \cite{CurveletDic} and contourlets \cite{ContourletDic}. Alternatively, it can be learned adaptively from input examples, a task referred to as ``dictionary learning'' (DL). The latter approach has provided state-of-the-art results in classic image processing applications, such as denoising \cite{KSVDDenoising}, inpainting \cite{KSVDInpainting}, demosaicing \cite{SparseApplications}, compression \cite{KSVDCompression,KSVDCompression2} and more.
Popular algorithms for dictionary learning are the MOD \cite{MOD} and the K-SVD \cite{KSVD}, which generalizes K-means clustering and learns an overcomplete dictionary that best sparsifies the input data.

Although successful in signal processing applications, the K-SVD algorithm ``as-is'' may not be suited for machine learning tasks such as classification or regression, as its primary goal is to achieve the best reconstruction of the input data, ignoring any discriminative information such as labels or annotations.
Many suggestions have been made to extend DL to deal with labeled data. The SRC method by Wright \textit{et al.} \cite{Classification} achieved impressive results in face recognition by sparse coding each test sample over a dictionary containing the train samples from all classes, and choosing the class that presents the best reconstruction error. In \cite{SupevisedDic1,SupevisedDic2} Mairal \textit{et al.} added a discriminative term to the DL model, and later incorporated the learning of the classifier parameters within the optimization of DL. The work reported in \cite{DKSVD} by Zhang \textit{et al.} was the first to incorporate the learning of the classifier parameters within the framework of the K-SVD algorithm. A similar extension has been made in \cite{LCKSVD,LCKSVD2} by Jiang \textit{et al.}, where in a addition to the classifier parameters, another discriminative term for the sparse codes was added and optimized using the regular K-SVD. In \cite{FDDL} Yang \textit{et al.} created an optimization function which forces both the learned dictionary and the resulting sparse coefficients to be discriminative. These algorithms and others that relate to them have been shown to be quite competitive with the best available learning algorithms, leading often times to state-of-the-art results.

In machine learning, kernels have provided a straightforward way of extending a given algorithm to deal with nonlinearities. Prominent examples of such algorithms include kernel-SVM \cite{KSVM}, kernel-PCA (KPCA) \cite{KPCA} and Kernel Fisher Discriminant (KFD) \cite{KFD}. Suppose the original data can be mapped to a higher dimensional ``feature space'', where tasks such as classification and regression are far easier. Under the proper conditions, the ``kernel trick'' allows one to train a learning algorithm in the higher-dimensional feature space, without using explicitly the exact mapping. This can be done by posing the entire algorithm in terms of inner products between the input signals, and later replacing these inner-products with kernels. One fundamental problem when using the kernel trick is that one is forced to access only the inner products of signals in feature space, instead of the signals themselves. A direct consequence of this is the need to store and manipulate a large kernel matrix $\bK$ of dimension $N \times N$ ($N$ being the size of the training set), which contains the modified inner products of all pairs of input examples.

In recent years, kernels have also been incorporated in the field of sparse representations, both in tasks of sparse coding \cite{KMP,KBP,KSR,KernelOMP,KernelSRC1,KernelSRC2,KernelSRC3} and dictionary learning \cite{KDL,KSR,SPSDKDL,KDL2,KernelizedDL,KDL3}. The starting point of this paper is the kernel DL method termed ``Kernel K-SVD'' (KKSVD) by Nguyen \textit{et al}.
The novelty in \cite{KDL} is in the ability to fully pose the entire DL scheme in terms of kernels, using a unique-structured dictionary which is a multiplication of two parts. The first, a constant matrix called the ``base-dictionary'', contains all of the mapped signals in feature space, and the second, called the ``coefficient-dictionary'', which is actually updated during the learning process.
The KKSVD suffers from the same issues arising when applying the kernel trick in general. Specifically, in large-scale datasets, where the number of input samples is of the order of thousands and beyond, the KKSVD quickly becomes impractical, both due to runtime and in the required storage space.

While kernel sparse representation is becoming more common, the existing algorithms are still challenging as they suffer from problems mentioned above. The arena of linear DL on the other hand, has a vast selection of existing tools that are implemented efficiently, enabling learning a dictionary quite rapidly in various settings and even if the number of examples to train on goes to the Millions. Indeed, in such extreme cases, online learning becomes appealing \cite{OnlineDic,OnlineDic2}.

As we show hereafter, our proposed method, ``Linearized Kernel Dictionary Learning'' (LKDL), enjoys the benefits of both worlds.
LKDL is composed of two stages: kernel matrix approximation, followed by a linearization of the training process by the creation of ``virtual samples'' \cite{LinearizedKSVM}. In the first stage, we apply the Nystr\"{o}m method to approximate the kernel matrix $\bK$, using a sub-sampled set of its columns. We explore and compare several such sub-sampling strategies, including core-sets, k-means, uniform, column-norm and diagonal sampling. Rather than using $\bK$ (or its approximation), we proceed with the assumption that it originates from a linear kernel, i.e. $\bK=\bF^T\bF$, and thus, instead of referring to $\bK$, we calculate the virtual samples $\bF$, using standard eigen-decomposition. After obtaining these virtual training and test sets, we apply an efficient off-the-shelf version of linear dictionary learning and continue with a standard classification scheme. This process essentially ``linearizes'' the kernel matrix and combines the nonlinear kernel information within that of the virtual samples.

We evaluate the performance of LKDL in three aspects: (1) first, we assure that the added nonlinearity in the form of the virtual datasets indeed improves classification results (with relation to linear DL) and performs comparably well as the exact kernelization performed in KKSVD; (2) we demonstrate the differences in runtime between the two methods and (3) we show the easiness of integration of LKDL with \textit{\textbf{any}} existing DL algorithm, including supervised DL.

We should note that a shorter version of this paper has been submitted to NIPS 2015. This paper extends over that submission in several ways: (i) it broadens the survey  of past work on supervised and kernel DL; (ii) it adds the combination of the proposed scheme with supervised DL, applied to two leading algorithms; and (iii) it expands the experimental results section substantially.

This paper is organized as follows: section \ref{linearDL} provides background to classical reconstructive DL with emphasis on the K-SVD and two methods of supervised DL, all of which are used later in the experimental part as the linear foundations over which our scheme is employed.  Section \ref{KDL} discusses Nguyen's KKSVD algorithm for kernel DL and discusses its complexity. Section \ref{LKDL} presents the details of our proposed algorithm, LKDL, for kernel DL. This section also builds a wider picture of this field, by surveying the relevant literature of incorporating kernels into sparse coding and the dictionary-learning.
Section \ref{Results} shows results corroborating the effectiveness of our method, and finally, section \ref{Conclusion} concludes this paper and proposes future research directions.

\section{Linear Dictionary Learning} \label{linearDL}

This section provides background on classic reconstructive DL, as well two examples of discriminative, supervised DL. The purpose of this section is to recall several key algorithms, the MOD and K-SVD, the FDDL, and the LC-KSVD, which we will kernelize in later sections.

\subsection{Background} \label{SS:Background}

In sparse representations, given an input signal $\bx \in \mathbb{R}^p$ and a ``dictionary'' $\bD \in \mathbb{R}^{p \times m}$, one wishes to find a ``sparse representation'' vector, $\bgamma \in \mathbb{R}^m$ such that $\bx \approx \tilde{\bx} = \bD \bgamma$. The dictionary $\bD = \left[\bd_1,\ldots,\bd_m\right]$ consists of ``atoms'' which faithfully represent the set of signals $\bx \in \cX$. The task of finding a signal's sparse representation is termed ``sparse coding''\footnote{The term ``Sparse Coding'' might be confusing because it is used in machine learning and brain research for describing the process we refer to as ``Dictionary Learning''. In this paper we follow the terminology of signal and image processing, and thus ``sparse-coding'' implies the quest for the sparse solution for an approximate linear system.} or ``atom decomposition'' and can be solved using the following optimization problem:
\begin{equation}\label{eq:SparseCoding}
    \bgamma = \underset{\bgamma}{\text{argmin}} \| \bx - \bD \bgamma \|_2^2 \quad s.t. \quad \| \bgamma \|_0 \leq \mathit{q},
\end{equation}
where $q$ is the number of nonzero coefficients in $\bgamma$, often referred to as the ``cardinality'' of the representation, and the term $\|\bgamma\|_0$ is the $l_0$-norm which counts the number of non-zeros in $\bgamma$. This problem is known to be NP-hard in general, implying that even for moderate $m$ (number of atoms), the amount of required computations becomes prohibitive. The group of algorithms which attempt to find an approximated solution to this problem are termed ``pursuit algorithms'', and they can be roughly divided into two main approaches.
The first are relaxation-based methods, such as the ``basis-pursuit'' \cite{basePursuit}, which relaxes the norm to be $l_1$ instead of $l_0$. The $l_1$-norm still promotes sparsity while making the optimization problem solvable with polynomial-time methods. The second family of algorithms used to approximate the solution of (\ref{eq:SparseCoding}) are the greedy methods, such as the ``matching-pursuit'' \cite{matchPursuit}, which find an approximation one atom at a time. In this paper we shall mostly address the latter group of pursuit algorithms, and more specifically, the Orthogonal Matching Pursuit (OMP) \cite{OMP} algorithm, which is known to be efficient and easy to implement.

\subsection{Classic Dictionary Learning} \label{SS:classicDL}

In ``dictionary learning'' (DL), one attempts to compute the dictionary $\bD \in \mathbb{R}^{p \times m}$ that best sparsifies a set of examples, serving as the input data $\bX \in \mathbb{R}^{p \times N}$. A commonly used formulation for DL is the following optimization problem:
\begin{equation}\label{eq:DictionaryLearning}
    \underset{\bD,\bGamma}{\text{argmin}} \| \bX - \bD \bGamma \|_F^2 \quad s.t. \quad 1 \le i\le N \quad
    \| \bgamma_i \|_0   \leq  \mathit{q},
\end{equation}
where $||\cdot||_F$ is the Frobenius norm and $\bGamma = \left[\bgamma_1,\ldots,\bgamma_N\right] \in \mathbb{R}^{m \times N}$ is a matrix containing the sparse coefficient vectors of all the input signals. The problem of DL can be solved iteratively using a Block Coordinate Descent (BCR) approach, of alternating between the sparse coding and dictionary update stages. Two such popular methods for DL are the MOD \cite{MOD} and K-SVD \cite{KSVD}.

In MOD \cite{MOD}, once the sparse coefficients in iteration $t$, $\bGamma_t$, are calculated using a standard pursuit algorithm, the optimization problem becomes:
\begin{equation}\label{eq:MOD1}
\bD_{t}=\underset{\bD}{\text{argmin}} \| \bX - \bD \bGamma_t \|_F^2.
\end{equation}
This convex sub-problem leads to the analytical batch update of the dictionary using Least-Squares:
\begin{equation}\label{eq:MOD2}
\bD_{t} = \bX \bGamma_t^T (\bGamma_t \bGamma_t^T)^{-1} = \bX \bGamma_t^{\dagger}.
\end{equation}
The problem with MOD is the need to compute the pseudo-inverse of the often very-large $\bGamma$. The K-SVD algorithm by Aharon \textit{et al.} \cite{KSVD} proposed alleviating this and speeding up the overall convergence by updating the dictionary one atom at a time.
This amounts to the use of the standard SVD decomposition of rank-$1$ for the update of each atom.

\subsection{Fisher Discriminant Dictionary Learning (FDDL)}

The work reported in \cite{FDDL} proposes an elegant way of performing discriminative DL for the purpose of classification between $L$ classes by modifying and extending the objective function posed in (\ref{eq:DictionaryLearning}). A fundamental feature of this method is the assumption that the dictionary is divided into $L$ disjoint parts, each serving a different class.

Let $\bX = \left[\bX_1,\ldots,\bX_L\right] \in \mathbb{R}^{p \times N}$ be the input examples of the $L$ classes, where $\bX_i \in \mathbb{R}^{p \times n_i}$ are the examples of class $i$.
Denote $\bD=[\bD_1,\ldots,\bD_L]\in \mathbb{R}^{p \times M}$ and $\bGamma=[\bGamma_1,\ldots,\bGamma_L] \in \mathbb{R}^{M \times N}$ the dictionary and the corresponding sparse representations.
The part $\bGamma_i \in \mathbb{R}^{M \times n_i}$ can be further decomposed as follows: $\bGamma_i=[(\bGamma_i^1)^T,\ldots,(\bGamma_i^j)^T,\ldots,(\bGamma_i^L)^T]^T$, where $\bGamma_i^j \in \mathbb{R}^{m_j \times n_i}$ are the coefficients of the samples $\bX_i \in \mathbb{R}^{p \times n_i}$ over the dictionary $\bD_j \in \mathbb{R}^{p \times m_j}$.
Armed with the above notations, we now turn to describe the objective function proposed in [14] for the discriminative DL task. This objective is composed of two parts. The first
is based on the following expression:
\begin{equation} \label{eq:FDDLFirst}
\begin{split}
& r\left(\bX_i,\bD,\bGamma_i\right) = \\
& \|\bX_i - \bD \bGamma_i\|_F^2 + \|\bX_i - \bD_i \bGamma_i^i\|_F^2 + \sum_{j=1 \atop j \neq i}^L {\|\bD_j \bGamma_i^j\|_F^2}
\end{split}
\end{equation}
The first term demands a good representation of the $i$-th class samples using the whole dictionary, and the second term further demands a good representation for these examples using their own class' sub-dictionary. The third term is of different nature, forcing the $i$-th class examples to minimize their reliance on the other sub-dictionaries. Naturally, the overall penalty function will sum the expression in (\ref{eq:FDDLFirst}) for all the classes $i$.

We now turn to describe the second term in the objective function, which relies on the Fisher Discriminant Criterion \cite{FisherCriterion}. We define two scatter expressions, both applied to the representations. The first, $S_W(\bGamma)$ computes the within class spread, while the second, $S_B(\bGamma)$ computes the scatter between the classes:
\begin{equation}\label{eq:FDDLSecond}
\begin{split}
    & S_W(\bGamma) = \sum\nolimits_{i=1}^L{\sum\nolimits_{\bgamma_k \in \Gamma_i}{(\bgamma_k-\bmu_i)(\bgamma_k-\bmu_i)^T}} \\
    & S_B(\bGamma) = \sum\nolimits_{i=1}^L{n_i (\bmu_i-\bmu)(\bmu_i-\bmu)^T},
\end{split}
\end{equation}
and $\bmu,\bmu_i$ are the mean vectors of the learned sparse coefficient vectors, $\bGamma$ and $\bGamma_i$ correspondingly.  Naturally, we aim to minimize the first while maximizing the second.

The final FDDL model is defined by the following optimization expression:
\begin{equation}\label{eq:FDDLFinal}
\begin{aligned}
     J_{\left(\bD,\bGamma\right)}= \underset{(\bD,\bGamma)}{\text{argmin}} &\Bigl \{ \sum\nolimits_{i=1}^L{r(\bX_i,\bD,\bGamma_i)}+\lambda_1 \|\bGamma\|_1 + \\
    & \lambda_2 \left[\text{tr}\left(S_W(\bGamma)-S_B(\bGamma)\right) + \eta \|\bGamma\|_F^2 \right]\Bigr \}.
\end{aligned}
\end{equation}
The term $\|\bGamma\|_F^2$ serves as a regularization that ensures the convexity of (\ref{eq:FDDLSecond}).

The detailed optimization scheme of this rather complex expression is described in \cite{FDDL}, along with two classification schemes, a global and a local one, depending on the size of the input dataset.

\subsection{Label Consistent KSVD (LC-KSVD)} \label{SS:LCKSVD}

In \cite{LCKSVD,LCKSVD2}, an alternative discriminative DL approach is introduced, in which the learning of the dictionary, along with the parameters of the classifier itself, is performed simultaneously, leading to the scheme termed ``Label-Consistent K-SVD'' (LC-KSVD). These elements are combined in one optimization objective, which is handled using the standard K-SVD algorithm.


In order to improve the performance of a linear classifier, an extra term is added to the reconstructive DL optimization function:
\begin{equation}\label{eq:LC-KSVD1}
\underset{\bD,\bT,\bGamma}{\text{argmin}} \|\bX - \bD \bGamma\|_F^2
  + \alpha \|\bQ - \bT \bGamma\|_F^2 \quad s.t \ \forall i, \, \|\bgamma_i\|_0 \leq q.
\end{equation}
The second term encourages the sparse coefficients to be discriminative. More specifically, the matrix $\bQ=[\bq_1,\ldots,\bq_N] \in \mathbb{R}^{m \times N}$ stands for the ``ideal'' sparse-coefficient matrix for discrimination, where $\bq_{i}$ is a binary vector encoding the assignment of each example to its destination atoms. The matrix $\bT \in \mathbb{R}^{m \times m}$ transforms the sparse codes $\bGamma$ to their idealized versions in $\bQ$. This term thus promotes identical sparse codes for input signals from the same class and orthogonal sparse codes for signals from different classes.


In addition to the discriminative term added above, the authors in \cite{LCKSVD} propose learning the linear classifier within the framework of the DL. A linear predictive classifier is used of the form: $f(\bgamma,\bTheta)=\bTheta \bgamma$, where $\bTheta \in \mathbb{R}^{L \times m}$. The overall objective function suggested is:
\begin{equation} \label{eq:LC-KSVD2}
\begin{split}
&\underset{\bD,\bTheta,\bT,\bGamma}{\text{argmin}} \Bigl \{ \|\bX - \bD \bGamma\|_F^2 + \alpha \|\bQ-\bT \bGamma\|_F^2\\
&  + \beta \|\bH - \bTheta \bGamma\|_F^2 \Bigr \}, \quad s.t. \quad \forall i, \ \|\bgamma_i\|_0 \leq q,
\end{split}
\end{equation}
where the classification error is represented by the term $\|\bH-\bTheta \bGamma\|_2^2$, $\bTheta$ contains the classifier parameters, $\bH = [\bh_1,\ldots,\bh_N] \in \mathbb{R}^{L \times N}$ is the label matrix of all input examples, in which the vector $\bh_i = [0,0,\ldots,0,1,0,\ldots,0]^T$ contains only zeros apart from the index corresponding to the class of the example. The optimization function in (\ref{eq:LC-KSVD2}) can also be written as follows:
\begin{equation}\label{eq:normalLC-KSVD2}
\underset{\bD_{new},\bGamma}{\text{argmin}} \|\bX_{new} - \bD_{new} \bGamma\|_F^2, \quad s.t. \quad \forall i, \quad \|\bgamma_i\|_0 \leq q,
\end{equation}
where $\bX_{new}=\left(\bX^T,\sqrt{\alpha}\bQ^T,\sqrt{\beta}\bH^T\right)^T \in \mathbb{R}^{(p+m+L) \times N}$ and $\bD_{new}=\left(\bD^T,\sqrt{\alpha}\bT^T,\sqrt{\beta}\bTheta^T\right)^T \in \mathbb{R}^{(p+m+L) \times m}$. The unified columns in $\bD_{new}$ are all normalized to unit $l_2$ norm.
The optimization objective in (\ref{eq:normalLC-KSVD2}) can be solved using standard DL algorithms, such as K-SVD.

The authors propose two cases of LC-KSVD: LC-KSVD2, in which the parameters of the classifier are learned along with the dictionary, as shown in (\ref{eq:LC-KSVD2}) and the second, LC-KSVD1, in which they are calculated separately by: $\bTheta = \left(\bGamma \bGamma^T + \tau_2 \bI\right)^{-1}\bGamma \bH^T$. More details on these expressions and the numerical scheme for minimizing the objective function can be found in \cite{LCKSVD,LCKSVD2}. A new sample $\bx$ is classified by first sparse coding over the dictionary $\hat{\bD}$, and then, applying the classifier $\hat{\bTheta}$ to estimate the label $j$.

\section{Kernel Dictionary Learning}\label{KDL}

This section focuses on kernel sparse representations, with emphasis of the kernel-KSVD method by Nguyen \textit{et al.}, which we will compare with later on this paper.

\subsection{Kernels - The Basics} \label{SS:Kernels}

In machine learning, it is well-known that a non-linear mapping of the signal of interest to higher dimension may improve its discriminability in tasks such as classification. Let $\bx \in \cX$ be a signal in input space, which is embedded to a higher dimensional space $\cF$ using the mapping $\Phi, \bx \in \mathbb{R}^{p} \rightarrow \Phi(\bx) \in \mathbb{R}^{P}$ ($P \gg p$ and it might even be infinite). The space in which this new signal $\Phi(\bx)$ lies is called the ``feature space''. The next step in machine learning algorithms, in particular in classification, would be learning a classifier based on the mapped input signals and labels. This task can be prohibitive if tackled directly. A way around this hurdle is the ``kernel trick'' \cite{RKHS,Kernels}, which allows computing inner products between pairs of signals in the feature space, using a simple nonlinear function operating on the two signals in input space:
\begin{equation}\label{eq:Kernel}
\kappa\left(\bx,\bx'\right)=\left<\Phi(\bx),\Phi(\bx')\right>=\Phi(\bx)^T\Phi(\bx'),
\end{equation}
where $\kappa$ is the ``kernel''. This relation holds true for positive-semi-definite (p.s.d) and Mercer kernels \cite{KSVM}. Thus, suppose that the learning algorithm can be fully posed in terms of inner products. In such a case, one can achieve a ``kernelized'' version by swapping the inner products with the kernel function, without ever operating in the feature space.

In case there are $N$ input signals $\bX = [\bx_1,\ldots,\bx_N] \in \mathbb{R}^{p \times N}$, the ``kernel matrix'' $\bK \in \mathbb{R}^{N \times N}$ holds the kernel values of all pairs of input signals:
\begin{equation}\label{eq:KernelMat}
\bK_{i,j} = \kappa(\bx_i,\bx_j) = \left<\Phi(\bx_i),\Phi(\bx_j)\right>, \quad \forall i,j = 1..N.
\end{equation}

An inherent constraint in kernel algorithms is the fact that the solution vectors, for example the principal components in KPCA, are expansions of the mapped signals in feature space:
\begin{equation}\label{eq:KPCA}
\bv = \sum_{i=1}^{N} \alpha_i\Phi(\bx_i).
\end{equation}
The subspace in which the possible solutions lie, can be viewed as an $N$ dimensional surface residing in $\cF$ \cite{InputFeature}. Motivated by the inability to directly approach the mapped signals in feature space, researchers have suggested embedding the $N$ dimensional surface to a finite Euclidean subspace, where all geometrical properties, such as distances and angles between pairs of $\Phi(\bx_i)'s$, are preserved \cite{Embedding}. The embedding is called the ``kernel empirical map'' and the resulting subspace is referred to as the ``empirical feature subspace''. One way to embed a given signal $\bx$ to the empirical feature space is by calculating kernel values originating from inner products with all input training examples:
\begin{equation}\label{eq:EmpiricalMap}
\bx \rightarrow \left[\kappa(\bx,\bx_1),\ldots,\kappa(\bx,\bx_N)\right]^T.
\end{equation}

\subsection{Kernel Dictionary Learning} \label{SS:KDL}

A straightforward way to kernelize dictionary learning would be exchanging all the signals (and dictionary atoms) with their respective representations in feature space: $\bx \rightarrow \Phi(\bx), \bd \rightarrow \Phi(\bd)$ and rephrasing the algorithm such that it contains solely inner products between pairs of these ingredients. A difficulty with this approach is that during the learning process, the dictionary atoms are in feature space. As there is no exact reverse mapping from the updated inner products to their corresponding signals in input space, there is no direct way of accessing the updated dictionary atoms, as practiced in linear DL.

In order to solve this problem, the authors in \cite{KDL} suggest decomposing the dictionary in feature space into: $\Phi(\bD) = \Phi(\bX) \bA$, where $\Phi(\bX)$ is the constant part, called the ``base-dictionary'', which consists of all mapped input signals, and $\bA$ is the only part updated during the learning, called the ``coefficient-dictionary''. Just like in the case of the KPCA \cite{KPCA}, the obtained  dictionary is limited to an $N$-dimensional manifold in the feature space.

The kernel dictionary learning can now be formulated as the following optimization problem:
\begin{equation}\label{eq:ChellapaKDL}
    \underset{\mathbf{A},\bGamma}{\text{argmin}} \| \Phi(\bX) - \Phi(\bX) \mathbf{A} \bGamma \|_F^2  \quad s.t. \quad \forall i=1..N \quad \| \bgamma_i \|_0 \leq \mathit{q}.
\end{equation}
Similarly to linear DL, this optimization problem can be solved iteratively by first performing sparse coding with a fixed dictionary $\bA$, then updating the dictionary according to the computed sparse representations $\bGamma$, and so on, until convergence is reached. The kernelized equivalent of sparse coding is given by: \begin{equation}\label{eq:KernelSparseCoding}
    \underset{\bgamma}{\text{argmin}} \| \Phi(\bz) - \Phi(\bX) \mathbf{A} \bgamma \|_2^2 \quad s.t. \quad \| \bgamma \|_0 \leq \mathit{q},
\end{equation}
where $\bz$ is the input signal. As mentioned earlier, the sparse coding algorithm we focus on in this paper, as well as in Nguyen's KKSVD \cite{KDL}, is the OMP \cite{OMP} and its kernel version, KOMP \cite{KDL}. Table \ref{table:kernelizingOMP} presents two of the main stages in the OMP algorithm, which are the Atom-Selection (AS) and Least-Squares (LS) stages, and their kernelized version. As can be seen, these stages can be completely represented using the coefficient dictionary $\bA$, the sparse representation vector $\bgamma$ and the kernel functions $\bK(\bX,\bX)\in \mathbb{R}^{N \times N}$ and $\bK(\bz,\bX) = [\kappa(\bz,\bx_1),\ldots,\kappa(\bz,\bx_N)] \in \mathbb{R}^{1 \times N}$.

The dictionary update stage, can also be kernelized. In the MOD algorithm \cite{MOD}, the update of $\bA$ in iteration $t+1$ is given by: $\bA_{t+1}=\bGamma_t^T(\bGamma_t\bGamma_t^T)^{-1}=\bGamma_t^{\dagger}$, being the solution to: $\underset{\bA}{\text{argmin}}\| \Phi(\bX) - \Phi(\bX)\bA\bGamma \|_F^2$. A similar update can be derived for the K-SVD algorithm, as described in depth in \cite{KDL,KDL2}.

\begin{table*}[!t]
\caption{Complexity of the atom selection (AS) and the least square (LS) stages in linear and kernel-OMP. $\bI_S$ is the current support vector and $|\bI_S|$ its length, $\bD_S$, $\bA_S$ and $\bgamma_S$ are sub-matrices of $\bD$, $\bA$, and $\bgamma$, respectively, corresponding to $\bI_S$. $\br_t$ is the residual.}
\label{table:kernelizingOMP}
\centering
\setlength{\extrarowheight}{5pt}
\begin{tabular}{|| c || c || c ||}
\multicolumn{1}{c}{} &\multicolumn{1}{c}{\bf Term}  &\multicolumn{1}{c}{\bf Complexity} \\
 \hline
 OMP-AS &
 $\left<\br_t,\bd_j\right>=\left<\bz-\bD_S\bgamma_S,\bd_j\right>=\bz^T\bd_j - \bgamma_S^T\bD_S^T\bd_j$ &
 $O\left(p|\bI_S| + p\right)$\\
 \hline

 KOMP-AS \cite{KDL} &
 $\bK(\bz,\bX)\ba_j - \bgamma_S^T\bA_S^T\bK(\bX,\bX)\ba_j$ &
 $O\left(N^2 + |\bI_S|N + N\right)$ \\
 \hline

 OMP-LS &
 $\bgamma_S = \left(\bD_S^T \bD_S\right)^{-1}\bD_S^T \bz$ &
 $O\left(p|\bI_S|^2 + p|\bI_S| + |\bI_S|^3\right)$\\
 \hline

 KOMP-LS \cite{KDL}&
 $\bgamma_S = \left[\bA_S^T \bK(\bX,\bX) \bA_S\right]^{-1}(\bK(\bz,\bX)\bA_S)^T$ &
 $O\left(N^2|\bI_S| + N|\bI_S| + |\bI_S|^3\right)$ \\
 \hline
\end{tabular}
\end{table*}

\subsection{Difficulties in KDL}\label{SS:KDLDifficult}

There are a few difficulties that arise when dealing with kernels, and specifically in kernel dictionary learning.
In the input space, a signal $\bx \in \mathbb{R}^p$ can be described using its own $p$ features, while in feature space it is described by its relationship with \textit{all of the other $N$ input signals}. The runtime and memory complexity of a kernel learning algorithm changes accordingly and depends on the number of input signals, instead of on the dimension of the signals. This observation is also true for Nguyen's KDL where the kernel matrix $\bK$ is used during the sparse coding and dictionary update stages, and must be stored in full. In applications where the number of input samples is large, this dependency on the kernel matrix becomes prohibitive. In table \ref{table:kernelizingOMP}, one can see the complexity of the main stages in the KOMP algorithm and compare it to the linear OMP version. It is clear that both the atom-selection and the least-squares stages are governed quadratically on the size of the input dataset.

Another inherent difficulty in kernel methods is the need to tailor each algorithm such that it is formulated solely through inner products. This constraint creates complex and cumbersome expressions and is not always possible, as some steps in the algorithm may contain a mixture of the signals and their mapped version.

\section{The Proposed Algorithm} \label{LKDL}

Section \ref{linearDL} and \ref{KDL} gave some background to the task we address in this paper. We saw that kernelization of the DL task can be beneficial, but unfortunately, we also identified key difficulties this process is accompanied by. In this work we aim to propose a systematic and simple path for kernelizing existing dictionary learning algorithms, in a way that will avoid the problems mentioned above. More specifically, we desire to be able to kernelize any existing DL algorithm, be it unsupervised or supervised, and do so while being able to work on massive training sets without the need to compute, store, or manipulate the kernel matrix K. In this section we outline such a solution, by carefully describing its key ingredients.

\subsection{Kernel matrix approximation} \label{SS:KernelMatApprox}

Let $\bX \in \mathbb{R}^{p \times N}$ be the input signals and $\bK \in \mathbb{R}^{N \times N}$ their corresponding kernel matrix. We shall further assume that $\bK$ is of rank $r \le N$. As long as the kernel satisfies Mercer's conditions of positive-semi-definiteness it can be written as an inner product between mapped signals in feature space: $\bK_{i,j}=\left<\Phi(\bx_i),\Phi(\bx_j)\right>$. Assume, for the sake of the discussion here, that the kernel function applies a simple inner product, i.e.: $\bK_{i,j}=\left<\bbf_i,\bbf_j\right>=\bbf_i^T\bbf_j$, where $\bbf_i,\bbf_j$ are the feature vectors corresponding to $\bx_i$ and $\bx_j$, respectively. Thus, the kernel matrix would have the form: $\bK=\bF^T\bF=\Phi(\bX)^T\Phi(\bX)$, where $\bF$ is a matrix of size $r \times N$ ($r$ is the feature-space dimension, and we have assumed that it is smaller than $N$). One can refer to the vectors $\{\bbf_i\}_{i=1}^N$ in $\bF$ as ``Virtual Samples'' \cite{LinearizedKSVM}. This way, instead of  learning using the kernel matrix $\bK$, one could work on these virtual samples directly using a linear learning algorithm, leading to the same outcome. In the following, we will leverage on this insight.

The kernel matrix is generally symmetric and positive-semi-definite, and as such can be decomposed using eigen-decomposition as follows: $\bK=\bU \boldsymbol{\varLambda} \bU^T$, where $\boldsymbol{\varLambda} \in \mathbb{R}^{r \times r}$ is a diagonal matrix containing all of the nonzero eigenvalues of $\bK$ in descending order and $\bU \in \mathbb{R}^{N \times r}$ contains the matching orthonormal eigenvectors. An approximation of the virtual samples can be achieved by:
\begin{equation}\label{eq:Mapping}
    \bF = \boldsymbol{\varLambda}^{1/2}  \bU^T = \boldsymbol{\varLambda}^{-1/2}  \bU^T  \bK.
\end{equation}
The virtual samples can be viewed as a mapping of the original input signals to an $r$-dimensional empirical feature space.
\begin{equation}\label{eq:VirtualSamples}
    \bx \rightarrow \boldsymbol{\varLambda}^{-1/2}  \bU^T  \left(\kappa(\bx,\bx_1),\kappa(\bx,\bx_2),\ldots,\kappa(\bx,\bx_N)\right)^T.
\end{equation}
An approximated kernel empirical map of dimension $k \leq r$ can also be obtained by considering only the top $k$ eigenvalues and corresponding eigenvectors $ \rightarrow \bF_k = \left( \boldsymbol{\varLambda}_k \right)^{1/2} \left( \mathbf{U}_k \right)^T$.

This ``linearization'' is the mediator between kernel DL which is obligated to store and manipulate the kernel matrix $\bK$, and linear DL that can deal with very large datasets.
The decomposition of the matrix $\bK$ to its eigenvalues and eigenvectors is a demanding task in itself, both in time $O(N^2k)$ and in space $O(N^2)$. Next we will show how a good approximation of the matrix $\bK$ can be constructed with only a subset of its columns, using the popular Nystr\"{o}m method.

\subsection{Nystr\"{o}m method} \label{SS:Nystrom}

A common necessity in many algorithms in signal processing and machine learning is deriving a relatively accurate and efficient approximation of a large matrix. An attractive method that has gained popularity in recent years is the Nystr\"{o}m method \cite{Nystrom,Nystrom2,Nystrom3}, which generates a low-rank approximation using a subset of the input data. The original Nystr\"{o}m method, first introduced by Williams and Seeger \cite{Nystrom}, proposed using uniform sampling without replacement.

Let $\bK \in \mathbb{R}^{N \times N}$ be a symmetric positive semi-definite matrix, and in particular for the discussion here, a kernel matrix. Suppose $c \leq N$ columns from the matrix $\bK$ are sampled uniformly without replacement to form the reduced matrix $\bC \in \mathbb{R}^{N \times c}$. Without loss of generality, the matrices $\bC$ and $\bK$ can be permuted as follows:
\begin{equation}\label{eq:NystromMatrixDecomp}
    \bC =
    \begin{bmatrix}
    \bW \\
    \bS
    \end{bmatrix}
    \quad and \quad
    \bK =
    \begin{bmatrix}
    \bW & \bS^T \\
    \bS & \bB
    \end{bmatrix},
\end{equation}
where $\bW \in \mathbb{R}^{c \times c}$ is the kernel matrix of the intersection of the chosen $c$ columns with $c$ rows, $\bB \in \mathbb{R}^{(N-c)\times (N-c)}$ is the kernel matrix composed of the $N-c$ remaining rows and columns, and $\bS \in \mathbb{R}^{(N-c) \times c}$, is a mixture of both.
The Nystr\"{o}m method uses both $\bC$ and $\bW$ to construct an approximation of the matrix $\bK$ as follows:
\begin{equation}\label{eq:Nystrom}
\bK \approx \bC \bW^{\dagger} \bC^T,
\end{equation}
where $(\cdot)^{\dagger}$ denotes the pseudo-inverse. The symmetric matrix $\bW$ can also be posed in terms of eigenvalues and eigenvectors: $\bW=\bV \boldsymbol{\Sigma} \bV^T$, where $\boldsymbol{\Sigma}$ is a diagonal matrix containing the eigenvalues of $\bW$ in descending order and $\bV$ contains the matching orthonormal eigenvectors. The pseudo-inverse of $\bW$ is given by $\bW^{\dagger}=\bV \boldsymbol{\Sigma}^{\dagger} \bV^T$. The expression of $(\bW^{\dagger})^{1/2}$ can be similarly derived: $(\bW^{\dagger})^{1/2}=(\boldsymbol{\Sigma}^{\dagger})^{1/2}\bV^T$.

We can represent $\bK$ as before, using linear inner-products of the virtual samples, and plug in Nystr\"{o}m's approximation:
\begin{equation}\label{eq:Nystrom2}
\bK = \bF^T \bF = \bC \bW^{\dagger} \bC^T = \bC \bV \boldsymbol{\Sigma}^{\dagger} \bV^T \bC^T,
\end{equation}
and derive the final expression of the virtual samples by:
\begin{equation}\label{eq:Nystrom3}
\bF = (\boldsymbol{\Sigma}^{\dagger})^{1/2} \bV^T \bC^T.
\end{equation}
The rank-$k$ $(k \leq c)$ approximation can similarly be derived:
\begin{equation}\label{eq:VirtualSamplesConcluded}
     \bF_k = \left(\boldsymbol{\Sigma}_k^{\dagger} \right)^{1/2} \mathbf{V}_k^T \bC^T,
\end{equation}
where $\boldsymbol{\Sigma_k}=\text{diag}(\sigma_1,\ldots,\sigma_k) \in \mathbb{R}^{k \times k}$ contains the $k$ largest eigenvalues of $\bW$ and $\bV_k \in \mathbb{R}^{c \times k}$, the corresponding orthonormal eigenvectors.

After performing the Nystr\"{o}m approximation, the space complexity of kernel DL reduces from $O(N^2)$ to $O(Nc)$, the size of the matrix $\bC$, which is used during the computation of the virtual samples. The time complexity of the Nystr\"{o}m method is $O(Nck+c^2k)$, where $O(Nck)$ represents the multiplication of $\mathbf{V}_k^T \bC^T$ and $O(c^2k)$ stands for the eigenvalue decomposition (and inversion) of the reduced matrix $\bW_k$.

Note that the process of computing the virtual samples may seem inefficient, but it is performed only once, after which the complexity of the DL is dictated by the chosen algorithm, and not by the ``kernelization''. In addition, in scenarios where the number of input examples is very large, the ratio $c/N$ in Nystr\"{o}m's method can be reduced greatly, i.e. $c \ll N$, making the approximation even less dominant in terms of runtime and memory, while retaining almost the same accuracy.

\subsection{Sampling Techniques} \label{SS:SamplingNystrom}
Since the Nystr\"{o}m method creates an approximation of a large symmetric matrix based on a subset of its columns, the chosen sampling scheme plays an important part. The basic method proposed originally by Williams and Seeger was uniform sampling without replacement \cite{Nystrom}. The columns of the Gram matrix can be alternatively sampled from a nonuniform distribution. Two such examples of nonuniform sampling include ``column-norm sampling'' \cite{ColumnNormSampling}, where the weight of the $i$th column $\bk^i$ is its $l_2$ norm: $p_i=\|\bk^i\|^2 / \|\bK\|_F^2$, and ``diagonal sampling'' \cite{DiagonalSampling} where the weight is proportional to the corresponding diagonal element: $p_i=\bK_{ii}^2 / \sum_{i=1}^N{\bK_{ii}^2}$. These methods can be made more sophisticated but require additional complexity: $O(N)$ in time and space for diagonal sampling and $O(N^2)$ for column-norm sampling. A comprehensive theoretical and empirical comparison of these three methods is provided in \cite{NystromSample}.

In \cite{KmeansSampling}, Zhang \textit{et al.} suggested an alternative approach of selecting a few ``representative'' columns in $\bK$ by first performing K-means clustering, then computing the reduced matrix $\bC$ based on these so-called ``cluster centers''. Denote by $\bX_R$ the resulting $c$ cluster centers, created from the original data $\bX$. The computation of the kernel matrices $\bC$ and $\bW$ would be: $\bC=\bK(\bX,\bX_R)$ and $\bW=\bK(\bX_R,\bX_R)$. Zhang \textit{et al.} also show that the combination of k-means clustering with the Nystr\"{o}m method minimizes the approximation error.

Another appealing sampling technique has been suggested in the context of coresets \cite{coreset}. The idea is to sample the given data by emphasizing unique samples that are ill-represented by the others. In the context of our problem, we sample $c$ signals from $\bX$ according to the following distribution: $p_i = err(\bx_i,\boldsymbol{\mu})/\sum_{\bx_i \in \bX} err(\bx_i,\boldsymbol{\mu})$, where $err(\bx_i,\boldsymbol{\mu})=||\bx_i - \boldsymbol{\mu} \gamma||_2^2$ is the representation error of the signal $\bx_i$, corresponding to the mean of all training signals $\boldsymbol{\mu}=(1/N)\sum_{i=1}^{N} \bx_i$.

\subsection{Linearized Kernel Dictionary Learning (LKDL)} \label{SS:Algorithm}

Let $\left\{\bx_i,y_i\right\}_{i=1}^N$ be a labeled\footnote{We consider here the case of labeled data, but the labels can be omitted, thus reducing to the simple representative DL format.} training set, arranged as a structure in $L$ categories: $\bX_{train}=\left[\bX_1,\ldots,\bX_L\right] \in \mathbb{R}^{p \times N}$, where $\bX_i$ contains the training samples that belong to the $i$th class and $N = \sum \nolimits_{i=1}^{L} n_i$. Our process of kernel dictionary learning is divided in two parts: the first, a pre-processing stage that creates new virtual training and test samples, followed by a second stage of applying a standard DL. This whole process is termed ``Linearized Kernel Dictionary Learning'' (LDKL).

The pre-processing stage is shown in algorithm \ref{LKDL-Preprocessing}. First, the initial training set $\bX_{train}$ is sampled using one of the techniques mentioned in section \ref{SS:SamplingNystrom}, creating the reduced set $\bX_R = [\bx_{R_1},\ldots,\bx_{R_c}] \in \mathbb{R}^{p \times c}$. Then the matrix $\bC \in \mathbb{R}^{N \times c}$ in Nsytr\"{o}m's method is calculated by simply applying the chosen kernel on each and every pair of columns in $\bX_{train}$ and $\bX_R$. Next, the reduced matrix $\bW \in \mathbb{R}^{c \times c}$ is both calculated and later on inverted using rank-$k$ eigen-decomposition.
Finally the virtual training samples $\bF_{train} \in \mathbb{R}^{k \times N}$ are calculated using equation (\ref{eq:VirtualSamplesConcluded}). The Nystr\"{o}m method permits approximating a new test vector $\bbf_{test}$ using equation (\ref{eq:VirtualSamples}), by using the mapping already calculated based on the training set, and multiplying by the joint kernel vector of the sampled set $\bX_R$ and the current test sample: $\bK(\bX_R,\bx_{test})$:
\begin{equation}\label{eq:MappingTest}
    \boldsymbol{f}_{test} = \left(\boldsymbol{\Sigma}_k^{\dagger} \right)^{1/2} \mathbf{V}_k^T \left[\kappa(\bx_{R_1},\bx_{test}),\ldots,\kappa(\bx_{R_c},\bx_{test}))\right]^T.
\end{equation}

Once the training and test sets are represented as virtual samples: $\bF_{train}$ and $\bF_{test}$, any linear DL-based classification method can be implemented. In the context of classification we follow Nguyen's ``distributive'' approach \cite{KDL2} of learning $L$ separate dictionaries $[\bD_1,\ldots,\bD_L]$ per each class, then classifying each test sample by first computing its sparse coefficient vector over each of the dictionaries $\{\bD_i\}_{i=1}^L$, and finally choosing the class corresponding to the smallest reconstruction error:
\begin{equation}\label{eq:classification}
r_i = \|\boldsymbol{f}_{test}-\bD_i \bgamma_i\|^2, \quad \forall i=1..L.
\end{equation}

\begin{algorithm}
\caption{LKDL Pre-Processing}\label{LKDL-Preprocessing}
\begin{algorithmic}[1]
\State \textbf{Input}: $\bX_{train}=\left[\bX_1,\ldots,\bX_L\right]$, $\bX_{test}$, the kernel $\kappa$, $smp\_method$, $c, k$
\State $\bX_R=sub\_sample(\bX_{train},smp\_method,c)$
\State Compute $\bC_{train}=\bK(\bX_{train},\bX_R)$
\State Compute $\bW=\bK(\bX_R,\bX_R)$
\State Approximate $\bW_k$ using $k$ largest eigenvalues and eigenvectors $\bW_k=\bV_k\boldsymbol{\Sigma}_k\bV_k^T$
\State Compute virtual train set $\bF_{train} = \left(\boldsymbol{\Sigma}_k^{\dagger} \right)^{1/2} \mathbf{V}_k^T \bC_{train}^T$
\State Compute $\bC_{test}=\bK(\bX_{test},\bX_R)$
\State Compute virtual test set $\bF_{test} = \left(\boldsymbol{\Sigma}_k^{\dagger} \right)^{1/2} \mathbf{V}_k^T \bC_{test}^T$
\State \textbf{Output}: $\bF_{train}=\left[\bF_1,\ldots,\bF_L\right]$, $\bF_{test}$
\end{algorithmic}
\end{algorithm}

\subsection{Relation to Past Work}\label{SS:literature}

The existing works on kernel sparse representations can be roughly divided to two categories.
The first corresponds to `analytical' methods that operate solely in the feature domain and use the kernel trick to find an analytical solution, be it sparse coding or dictionary update \cite{KernelOMP,KSR,KDL,KernelizedDL}. The other category refers to `empirical' or `approximal' methods that operate in the input space, while making some approximation or assumption regarding the mapped signals in feature space, in order to alleviate some of the constraints when working with kernels \cite{KMP,KBP,KernelSRC1}. Naturally, our work belongs to the second group of contributions.

In 2002, Bengio \textit{et al.} \cite{KMP} kernelized the matching pursuit algorithm by using the kernel empirical map of the input training examples as dictionary atoms. By referring to the kernel empirical map $\Phi_e$ instead of the actual mapped signals in $\cF$, the authors could perform standard linear matching pursuit without having to rewrite the algorithm in terms of inner products. In this case, the constraint of a p.s.d kernel was no longer mandatory. In 2005 \cite{KBP}, a similar concept of embedding the signals to a kernel empirical map was used to kernelize the basis pursuit algorithm. This approach of working in the input domain with an approximation of the kernel feature space is very similar to ours and can be described by the following embedding, evaluated over the entire training dataset $\{\bx_i\}_{i=1}^N$:
\begin{equation}\label{eq:kernelEmpiricalMap}
\bx \rightarrow \Phi_e(\bx) = \left[\kappa(\bx_1,\bx),\ldots,\kappa(\bx_N,\bx)\right]^T.
\end{equation}
The case in our algorithm, where all the training signals are involved in the approximation of the kernel matrix ($c=N,\bC=\bW=\bK$), results in a similar expression for the virtual samples:
\begin{equation}\label{eq:FullVirtaulSamples}
\bF=(\boldsymbol{\Sigma}^{1/2})^{\dagger} \bV^T \bC^T=(\boldsymbol{\Sigma}^{1/2})^{\dagger} \bV^T \bK^T,
\end{equation}
where $\boldsymbol{\Sigma}$ and $\bV$ are the eigenvalues and eigenvectors of the matrix $\bK$.
The embedding in this case is thus
\begin{equation}\label{eq:kernelEmpiricalMap2}
\Phi_e(\bx) = (\boldsymbol{\Sigma}^{1/2})^{\dagger} \bV^T \left[\kappa(\bx_1,\bx),\ldots,\kappa(\bx_N,\bx)\right]^T.
\end{equation}
Contrary to \cite{KMP,KBP}, our embedding preserves the similarities in the high-dimensional feature space, represented by the inner products, i.e,
\begin{equation}\label{eq:preserveInnerProduct}
\Phi_e(\bx)^T\Phi_e(\bx') \approx \kappa(\bx,\bx')=\Phi(\bx)^T\Phi(\bx'),
\end{equation}
where we have used the expression $\bK^{\dagger}=\bV \boldsymbol{\Sigma}^{\dagger} \bV^T$. In addition, both \cite{KMP} and \cite{KBP} focus on sparse coding only and do not address the accuracy of the kernel empirical map, nor its dimension, which can be highly restrictive in large-scale datasets.

Both Gao \textit{et al.} in 2010 \cite{KSR} and Li \textit{et al.} in 2011 \cite{KernelOMP}, proposed an analytical approach of kernelizing the basis pursuit and orthogonal matching pursuit algorithms. Contrary to \cite{KMP} and \cite{KBP}, the authors replaced all the inner products by kernels and worked entirely in the feature domain. Classification of faces and objects were achieved in \cite{KSR} using a similar approach as in the SRC algorithm \cite{Classification}.
Aside from kernelizing the SRC algorithm, \cite{KSR} also suggested updating the dictionary one atom at a time. By zeroing the derivative of the optimization function with respect to each atom, the authors acquired in the same term, a mixture of both the atom itself and its kernel with the input examples. As the resulting equation could not be solved analytically, an iterative fixed point update was implemented.

In 2012 Zhang \textit{et al.} \cite{KernelSRC1} provided an alternate approach of kernelizing the SRC algorithm. Instead of working with the implicit mapped signals in the feature space $\Phi(\by)$, the authors performed dimensionality reduction first, using the KPCA algorithm, then fed the resulting nonlinear features to a linear $l_1$ basis pursuit solver.
It can be shown that kernel PCA eventually entails the eigendecomposition of the kernel matrix (more accurately, the centered kernel matrix), as does our algorithm. The difference is that our method, apart from providing an accurate kernel mapping which preserves similarities in feature space, also avoids dealing with the kernel matrix altogether in the training stage, making it possible to work with large datasets.

\section{Experimental Results} \label{Results}

In the following section we highlight the three main benefits of incorporating LKDL with existing DL: (1) improvement in discriminability, which results in better classification (2) a small added computational effort by LKDL in comparison with typical kernel methods and (3) the ability to incorporate the LKDL seamlessly in virtually any existing linear DL algorithm, contributing to more compact dictionaries and sparse representations.

\subsection{Unsupervised Dictionary Learning} \label{SS:Unsupervised}

In this part we demonstrate the performance of our algorithm in digit classification on the USPS and MNIST databases. Our method of classification consists of first pre-processing the training and test data using LKDL, then performing regular, standard dictionary learning, using existing tools and finally deploying the classification scheme in section \ref{SS:Algorithm}. For sparse coding and dictionary learning, we use the batch-OMP and efficient-KSVD implementations from the latest OMP-Box (v10) and KSVD-Box (v13) libraries\footnote{Found in \url{http://www.cs.technion.ac.il/~ronrubin/software.html}} \cite{EfficientKSVD}.
During all experiments we use the KKSVD algorithm explained in section \ref{SS:KDL} \cite{KDL,KDL2} as our reference, in addition to regular linear KSVD. We use the original code of Nguyen's KKSVD\footnote{Found in \url{http://www.umiacs.umd.edu/~hien/KKSVD.zip}}. A fair comparison in accuracy and runtime, between LKDL and KKSVD can be made, as KKSVD uses the same functions from the OMP and KSVD libraries mentioned earlier. The k-means\footnote{K-means - \url{http://www.mathworks.com/matlabcentral/fileexchange/31274-fast-k-means/content/fkmeans.m}} and coreset\footnote{Coreset - \url{http://web.media.mit.edu/~michaf/index.html}} sampling techniques were also adopted from existing code. All of the tests were performed on a 64-Bit Windows7 Intel(R) Core(TM) i7-4790K CPU with 16GB memory.
The initial dictionary is a random subset of $m$ columns from the training set in each class.

\subsubsection{USPS dataset}\label{SSS:USPS}

The USPS dataset consists of 7,291 training and 2,007 test images of digits of size $16 \times 16$. All images are stacked as vectors of dimension $p=256$ and normalized to unit $l_2$ norm. Following the experiment in \cite{KDL2}, we choose the following parameters: 300 dictionary atoms per class, cardinality of 5 and 5 iterations of DL. The chosen kernel is polynomial of order 4, i.e. $\kappa(\bx,\bx')=(\bx^T\bx')^4$. The approximation parameters were chosen empirically using coarse-to-fine search and were set to: $c=20\%$ of $N$ training samples and $k=256$, the original dimension of the digits. The displayed results are an average of 10 repeated iterations with different initialization of the sub-dictionaries and different sampled columns $\bX_R$ in Nystr\"{o}m's method.

\begin{figure*}[!t]
\centering
\subfloat[]{\includegraphics[width=0.45\textwidth]{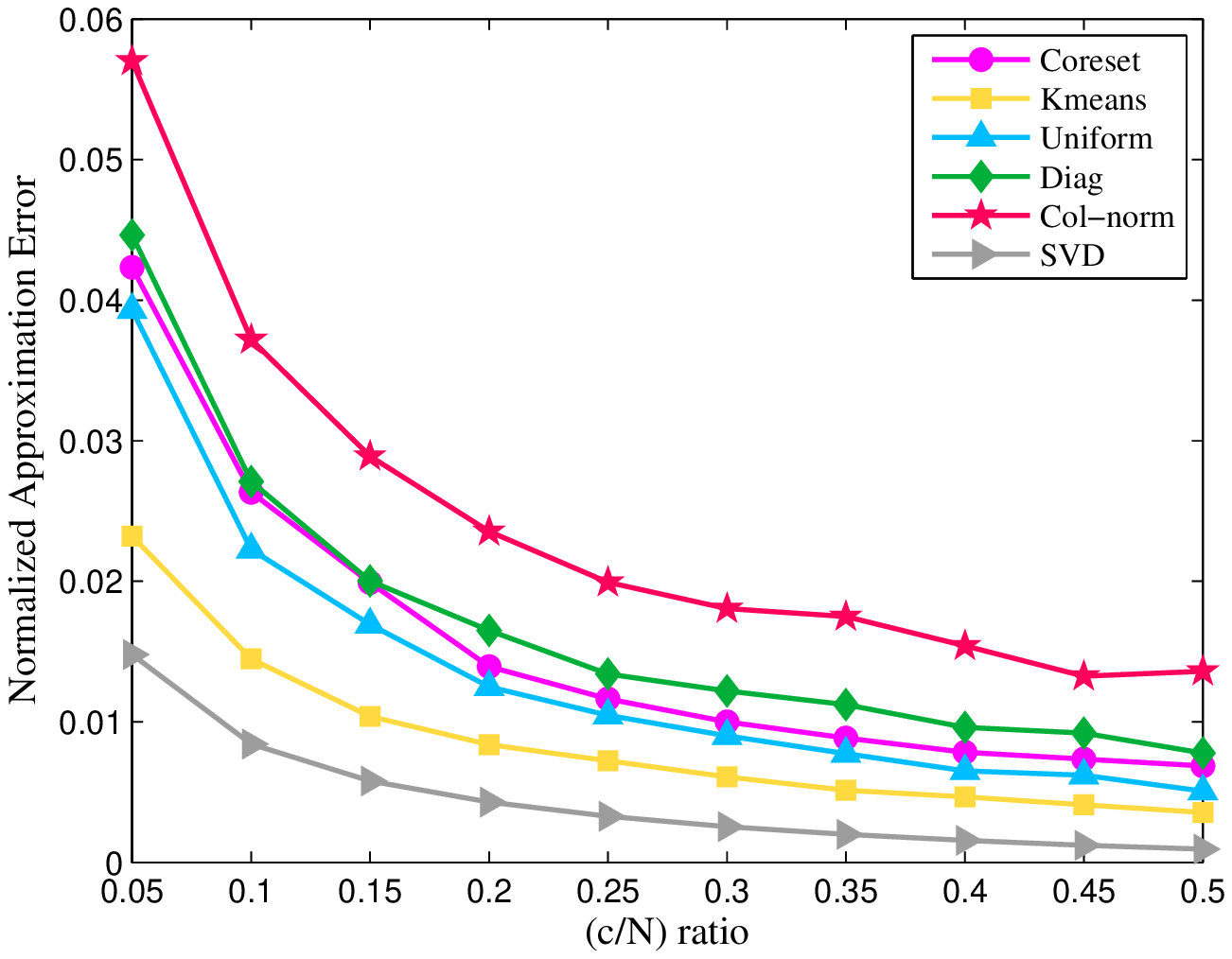}\label{fig:Approx_Error}}
\hfil
\subfloat[]{\includegraphics[width=0.45\textwidth]{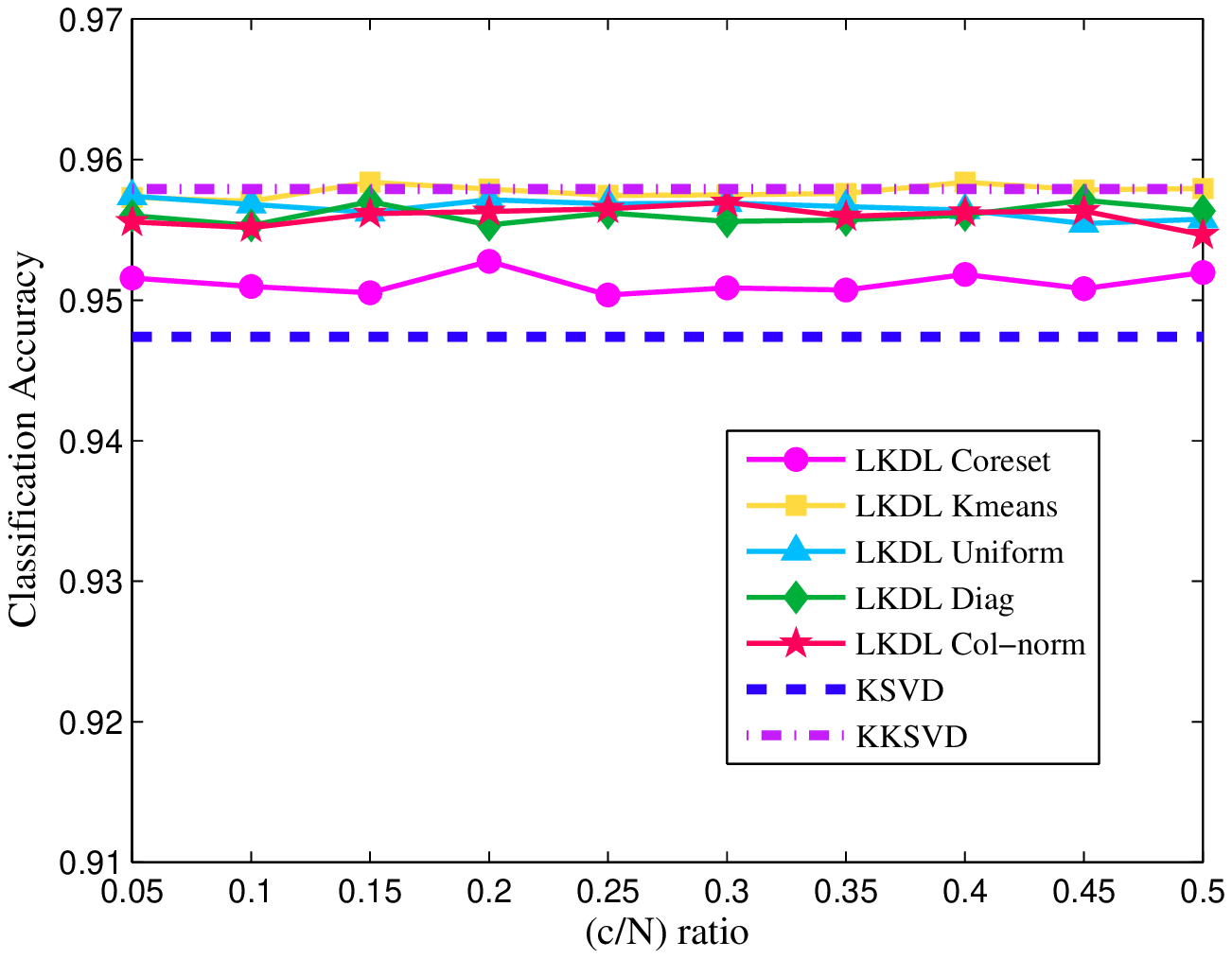}\label{fig:Approx_Accuracy}}
\caption{Approximation error (a) and classification accuracy (b) as a function of $c/N$, percent of samples used in Nystr\"{o}m method.}
\label{fig:GRAPH_1}
\end{figure*}

First we evaluate the quality of the representation of the kernel matrix using Nystr\"{o}m's method. We randomly choose 2,000 samples from USPS and approximate the resulting kernel matrix. In order to isolate the effect of column sub-sampling, we do not perform additional dimensionality reduction using eigen-decomposition and thus choose $k=256$. Five sampling techniques were examined: uniform \cite{Nystrom}, diagonal \cite{DiagonalSampling}, column-norm \cite{ColumnNormSampling}, k-means \cite{KmeansSampling} and coreset \cite{coreset}. We also added the ideal reconstruction using rank-$c$ SVD decomposition, which is optimal with respect to minimizing the approximation error, but takes much longer time to compute. We perform the comparison using the normalized approximation error:
\begin{equation}\label{eq:ApproxError}
    err = \frac{\|\bK -\widetilde{\bK}\|_F}{\|\bK\|_F},
\end{equation}
where $\bK$ is the original kernel matrix and $\widetilde{\bK}$ its Nystr\"{o}m approximation. Fig. \ref{fig:Approx_Error} shows the quality of the approximation versus the $c/N$ ratio, the percent of samples chosen for the Nystr\"{o}m approximation. As expected, SVD performs the best, as it is meant exactly for the purpose of providing the ideal rank-$c$ approximation of $\bK$. The second best approximation is obtained by k-means, which provides 98.5\% accuracy in terms of the normalized approximation error, with only 10\% of the samples. All other methods perform roughly the same. The differences in approximation quality reduce as the percent of chosen samples grows to half of the input dataset.

Next we examine the effect of sub-sampling on the classification accuracy of the entire database of USPS. Fig. \ref{fig:Approx_Accuracy} shows the classification accuracy as a function of $c/N$, along with the constant results of linear KSVD and KKSVD (which do not depend on $c$). There is a gap of $1\%$ between the results of linear KSVD and its kernel variants, which suggests that kernelization improves the discriminability of the input signals. It can be seen that k-means sampling again performs best and reaches classification accuracy of KKSVD, with only a fraction of the samples. In general, the percent of samples in Nystr\"{o}m approximation does not have much impact on the final classification accuracy (apart from small fluctuations that arise from the randomness of each run). This can be explained by the simplicity of the digit images and the relatively large number of training examples.

\begin{figure*}[!t]
\centering
\subfloat[]{\includegraphics[width=0.45\textwidth]{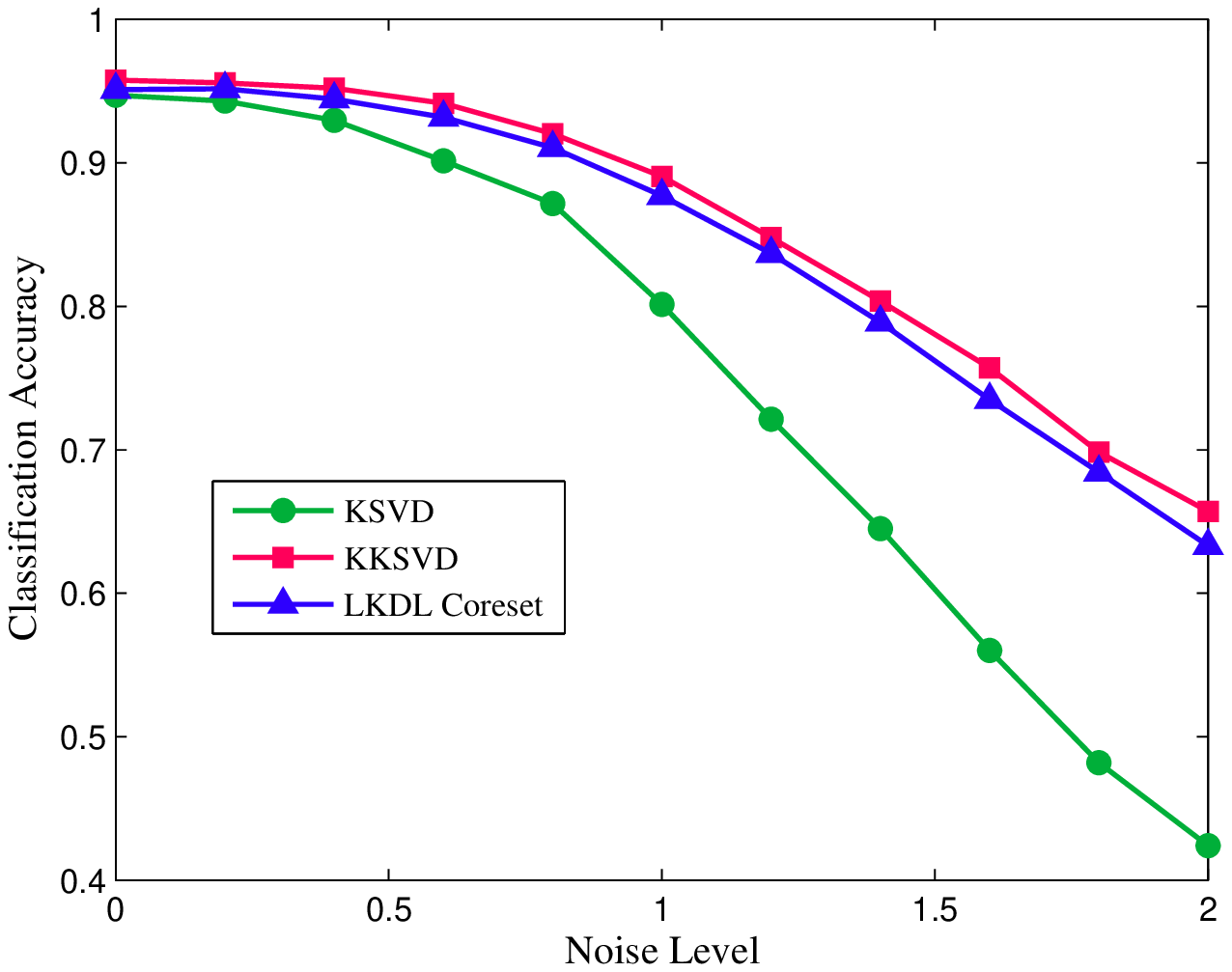}\label{fig:USPS_Noise}}
\hfil
\subfloat[]{\includegraphics[width=0.45\textwidth]{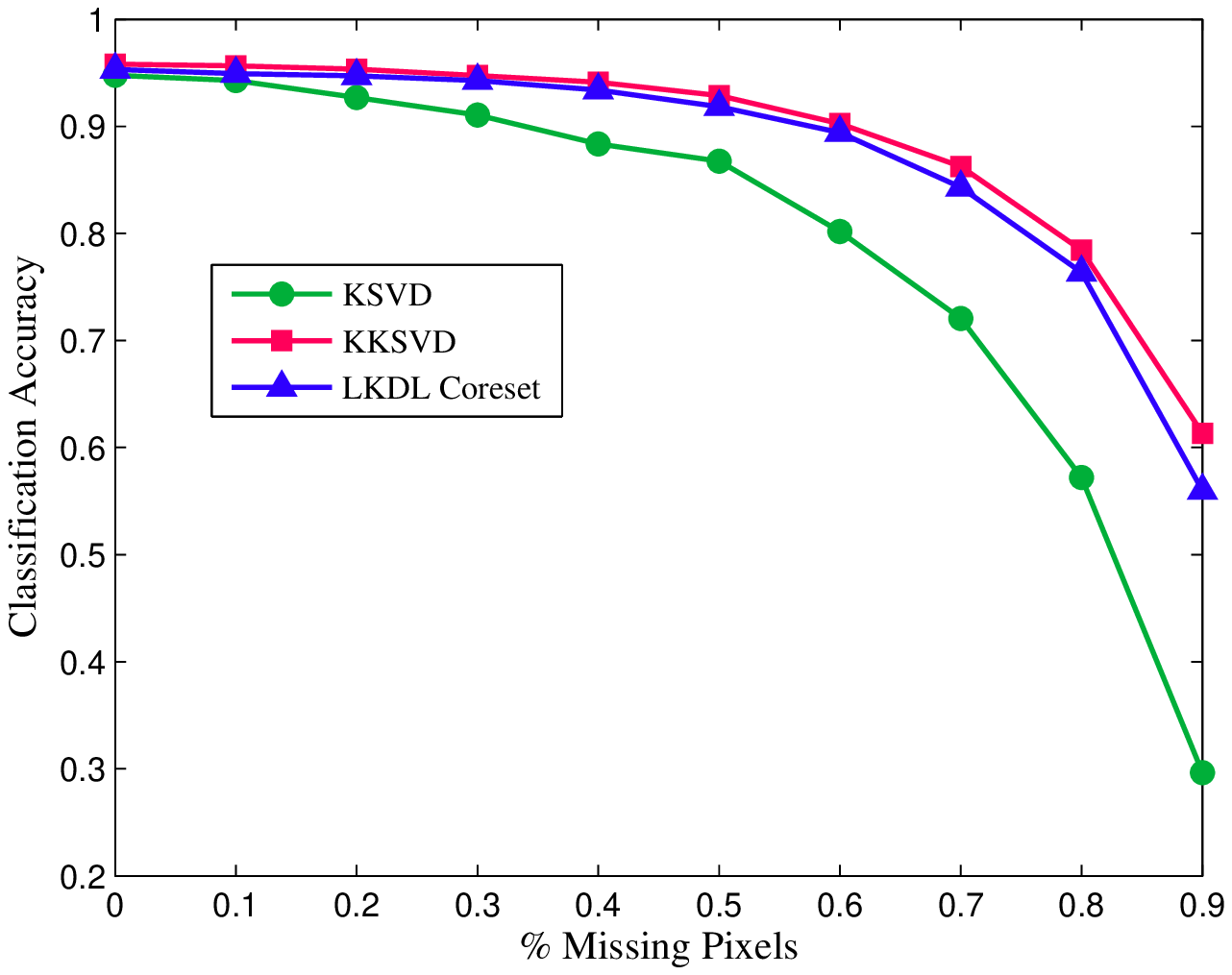}\label{fig:USPS_Pixels}}
\caption{Classification accuracy in the presence of Gaussian noise (a) and missing pixels (b).}
\label{fig:GRAPH_2}
\end{figure*}

Following Nguyen's setup in \cite{KDL} and \cite{KDL2}, we inspect the effect of corrupting the test images with white Gaussian noise and missing pixels. We use the same parameters as before and repeat the experiment 10 times with different random corruptions. The results of classification accuracy versus the standard deviation of the noise and the percent of missing pixels are given in Fig. \ref{fig:USPS_Noise} and \ref{fig:USPS_Pixels}. It is evident that adding the kernel improves the robustness of the database to both noise and missing pixels. The performance of LKDL follows that of KKSVD with a only 20\% of the training samples. The trend shown in our results is similar to that in \cite{KDL2}, although the results are slightly lower. This can be explained by the fact that in \cite{KDL2}, the authors did not use the traditional partitioning of training and test data of the USPS dataset. 
In this simulation, the coreset sampling technique was the best in dealing with signal corruptions, which is the reason it is the only method shown.

\subsubsection{MNIST dataset} \label{SS:MNIST Results}

Next we demonstrate the differences in runtime between our method and KKSVD using the larger-scale digit database of MNIST, which consists of 60,000 training and 10,000 test images of digits of size $28 \times 28$. Same as before, the digits were stacked in vectors of dimension $p=784$ and normalized to unit $l_2$ norm. We examine the influence of gradually increasing the training set on the classification accuracy and training time of the input data. In this simulation, the entire training set of 60,000 examples is reduced by randomly choosing a defined fraction of the samples, while maintaining the test set untouched. The runtime measured in LKDL includes the time needed to prepare both the training and test virtual samples, along with training the entire input dataset using linear KSVD. As for KKSVD, the runtime includes the preparation of the kernel sub-matrices for each class and the kernel DL using KKSVD. Parameters in the simulation were: 2 DL iterations, cardinality of 11, 700 atoms per digit, polynomial kernel of order 2, $c=15$\% and $k=784$. The results were averaged over 5 runs.

\begin{figure*}[!t]
\centering
\subfloat[]{\includegraphics[width=0.45\textwidth]{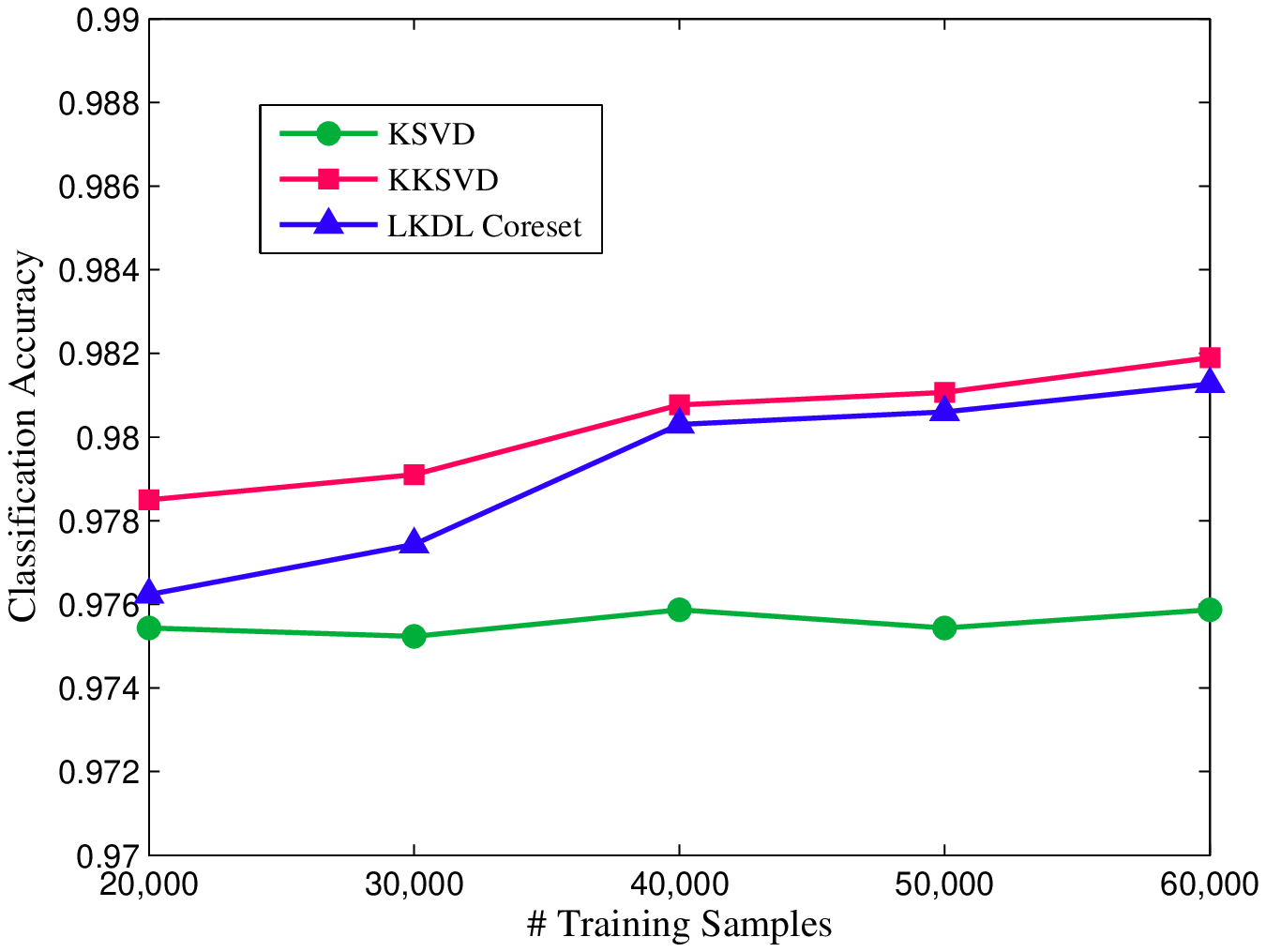}\label{fig:MNIST_accuracy}}
\hfil
\subfloat[]{\includegraphics[width=0.45\textwidth]{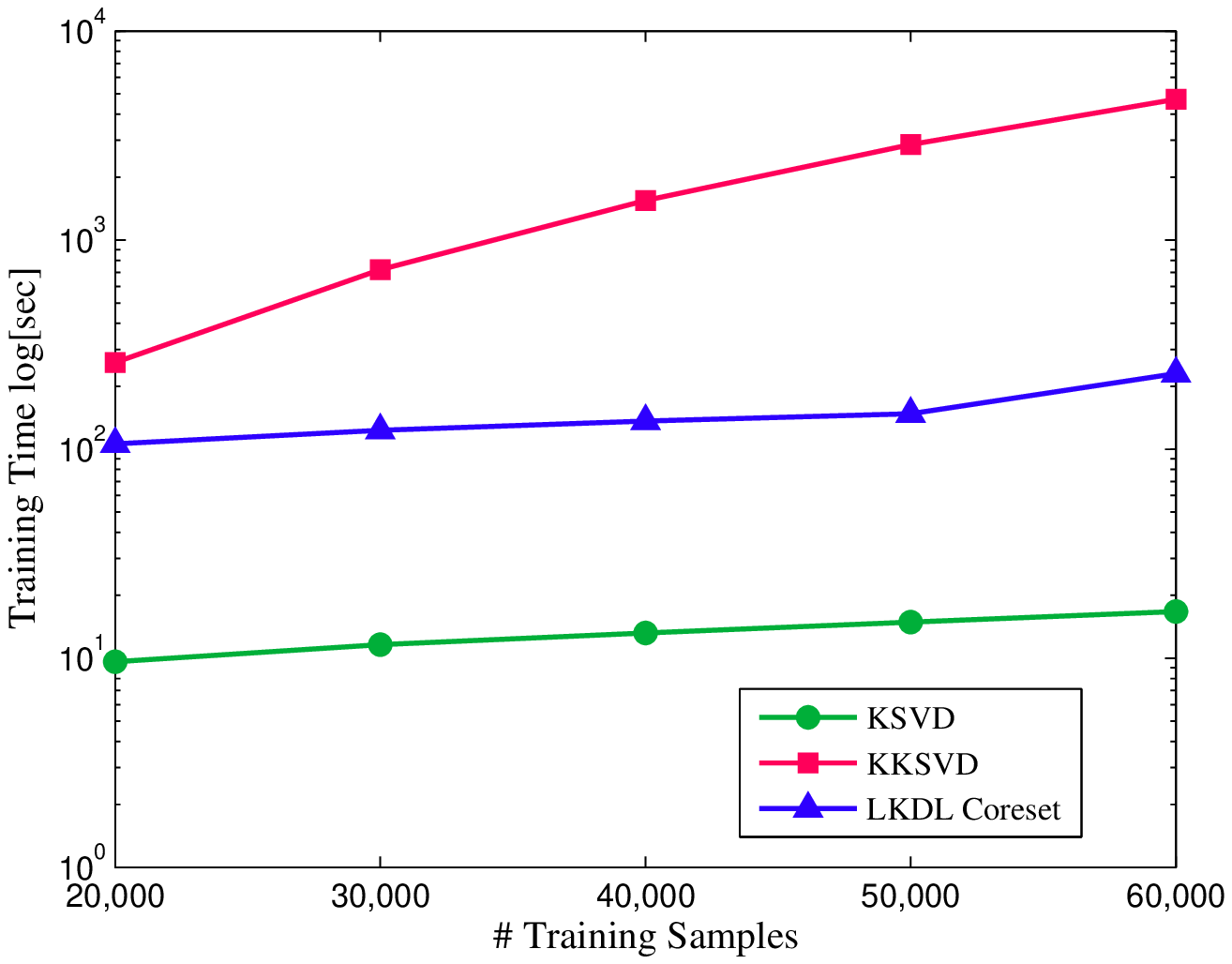}\label{fig:MNIST_runtime}}
\caption{Accuracy (a) and total training time (b) versus the number of input training examples in MNIST database. Runtime is shown in logarithmic scale.}
\label{fig:GRAPH_3}
\end{figure*}

The results can be seen in Fig. \ref{fig:MNIST_accuracy} and \ref{fig:MNIST_runtime}. Again, the coreset sampling method was chosen, as it provided the best results. The accuracy of LKDL versus KKSVD is comparable, while slightly worse, due to the approximation, but still better than the linear version of KSVD. The runtime of LKDL follows the one of KSVD, along with a component of calculating the virtual datasets. This is expected since our method ``piggy-backs'' on KSVD's performance and complexity. KKSVD's performance however, is dependent quadratically on the number of input samples in each class. When the database is large, the calculation of the virtual datasets (which is performed only once), is negligible versus the alternative of performing kernel sparse coding thousands of times during the DL process.

Note that we chose a relatively small number of DL iterations in order to reduce the already-long computation time of KKSVD. A larger number of DL iterations will lead to an even greater difference in runtime between KKSVD and LKDL.
For training the entire database of MNIST, LKDL is 19-times faster that KKSVD.


\subsection{Supervised Dictionary Learning}\label{SS:Supervised}

In the following set of experiments we demonstrate the easiness of combining our pre-processing stage with any DL algorithm, in particular the LC-KSVD \cite{LCKSVD} and FDDL \cite{FDDL}, both of which are supervised dictionary learning techniques that were mentioned earlier. We do so using the original code of LC-KSVD\footnote{Found in \url{http://www.umiacs.umd.edu/~zhuolin/LCKSVD/}} and FDDL\footnote{Found in \url{http://www.vision.ee.ethz.ch/~yangme/database_mat/FDDL.zip}}.
Throughout all tests, the training and test sets were pre-processed using LKDL to produce virtual training and test sets, which were later on fed as input to the DL and classification stages of each method. In all experiments, no code has been modified, except for exterior parameters which can be tuned to provide better results. The point in this setup is using an existing technique of supervised DL and showing the improvement that our method can provide.

\subsubsection{Evaluation on the USPS Database}

We start with comparing the classification accuracy of the same database from before, the USPS. First we perform regular FDDL with the following parameters: 5 DL iterations, 300 dictionary atoms per class, where the dictionary is first initialized using K-means clustering of the training examples. The scalars controlling the tradeoff in the DL and optimization expressions remained the same as in the demo provided by the authors: $\lambda_1=0.1,\lambda_2=0.001$ and $g_1=0.1,g_2=0.001$ (in \cite{FDDL}, these are referred to as $\gamma_1,\gamma_2$). As for LKDL pre-processing, the chosen parameters were: Polynomial kernel of degree 3, K-means based sub-sampling of 20\% of the training samples ($c/N=0.2$) and $k=256$. All results were averaged over 10 iterations with different initializations.

Table \ref{table:FDDL USPS_performance} shows the classification results with and without LKDL. There is a clear improvement in the results when adding LKDL as pre-processing. However the obtained results in this experiment are lower than those reported in \cite{FDDL}. This can be explained by the fact that we used the original database of USPS, while the provided code had a demo intended for an extended translation-invariant version of USPS. In addition, the exterior parameters $\lambda_1,\lambda_2,g_1,g_2$ were tweaked especially for the extended USPS, thus may have provided worse results in our case.

\begin{table}[!t]
\caption{Classification accuracy of FDDL on the USPS digit database, with and without LKDL pre-processing}
\label{table:FDDL USPS_performance}
\centering
\begin{tabular}{||l||c||}
\multicolumn{1}{l}{\bf Algorithm}  &\multicolumn{1}{c}{\bf Accuracy} \\
\hline
FDDL & 95.79 \\
\hline
FDDL + LKDL & \textbf{96.03} \\
\hline
\end{tabular}
\end{table}

\subsubsection{Evaluation on the Extended YaleB Database}

Next, we show the benefit of combining our method with LC-KSVD on the ``Extended YaleB'' face recognition database, which consists of 2,414 frontal images that were taken under varying lighting conditions. There are 38 classes in YaleB and each class roughly contains 64 images, which are split in half to training and test sets, following the experiment described in \cite{LCKSVD}. The original $192 \times 168$ images are projected to 504-dimensional vectors using a randomly generated constant matrix from a zero-mean normal distribution.
We use a dictionary size of 570 (in average 15 images per class) and sparsity factor of 30, same as in \cite{LCKSVD}. The kernel chosen for LKDL was Gaussian of the form: $\kappa(\bx,\bx')=\text{exp}\left(-\|\bx-\bx'\|_2^2/2{\sigma}^2\right)$, where $\sigma=1$. Due to the small size of the dataset, no sub-sampling was performed and $c$ was set to be the entire size of the training set. The value of the parameter $k$ (the dimension of the signal after eigen-decomposition) was set to 400, as it appeared that further dimensionality reduction of the already reduced 504-dimensional vector improved the results. In order to use the Gaussian kernel, the samples in the training and test sets were $l_2$ normalized, thus the original parameters of $\sqrt{\alpha}$ and $\sqrt{\beta}$ in expression (\ref{eq:LC-KSVD2}) had to be changed from 4 and 2 to $1/30$ and $1/91$ correspondingly. These parameters were chosen using a coarse-to-fine search and provided the best classification results. We use the original classification scheme in \cite{LCKSVD,LCKSVD2}.

Table \ref{table:LCKSVD YaleB_performance} shows the classification results of LC-KSVD1 and LC-KSVD2, with and without LKDL pre-processing. It is clear that the addition of the nonlinear kernel function increases the discriminability of the input samples and improves classification results by up to 1.8\% and 1.3\% in the case of LC-KSVD1 and LC-KSVD2, correspondingly. In fact, it appears that our LKDL blurs the differences between these two methods, meaning, there is no preference as to whether the classifier will be learned separately or jointly along with the dictionary.

\begin{table}[!t]
\caption{Classification accuracy of LC-KSVD1 and LC-KSVD2 on the Extended YaleB database, with and without LKDL pre-processing}
\label{table:LCKSVD YaleB_performance}
\centering
\begin{tabular}{||l||c||}
\multicolumn{1}{l}{\bf Algorithm}  &\multicolumn{1}{c}{\bf Accuracy} \\
\hline
LC-KSVD1 & 94.49 \\
\hline
LC-KSVD1 + LKDL & \textbf{96.33} \\
\hline
\hline
LC-KSVD2 & 94.99 \\
\hline
LC-KSVD2 + LKDL & \textbf{96.33} \\
\hline
\end{tabular}
\end{table}

\begin{figure*}[!t]
\centering
\subfloat[]{\includegraphics[width=0.25\textwidth]{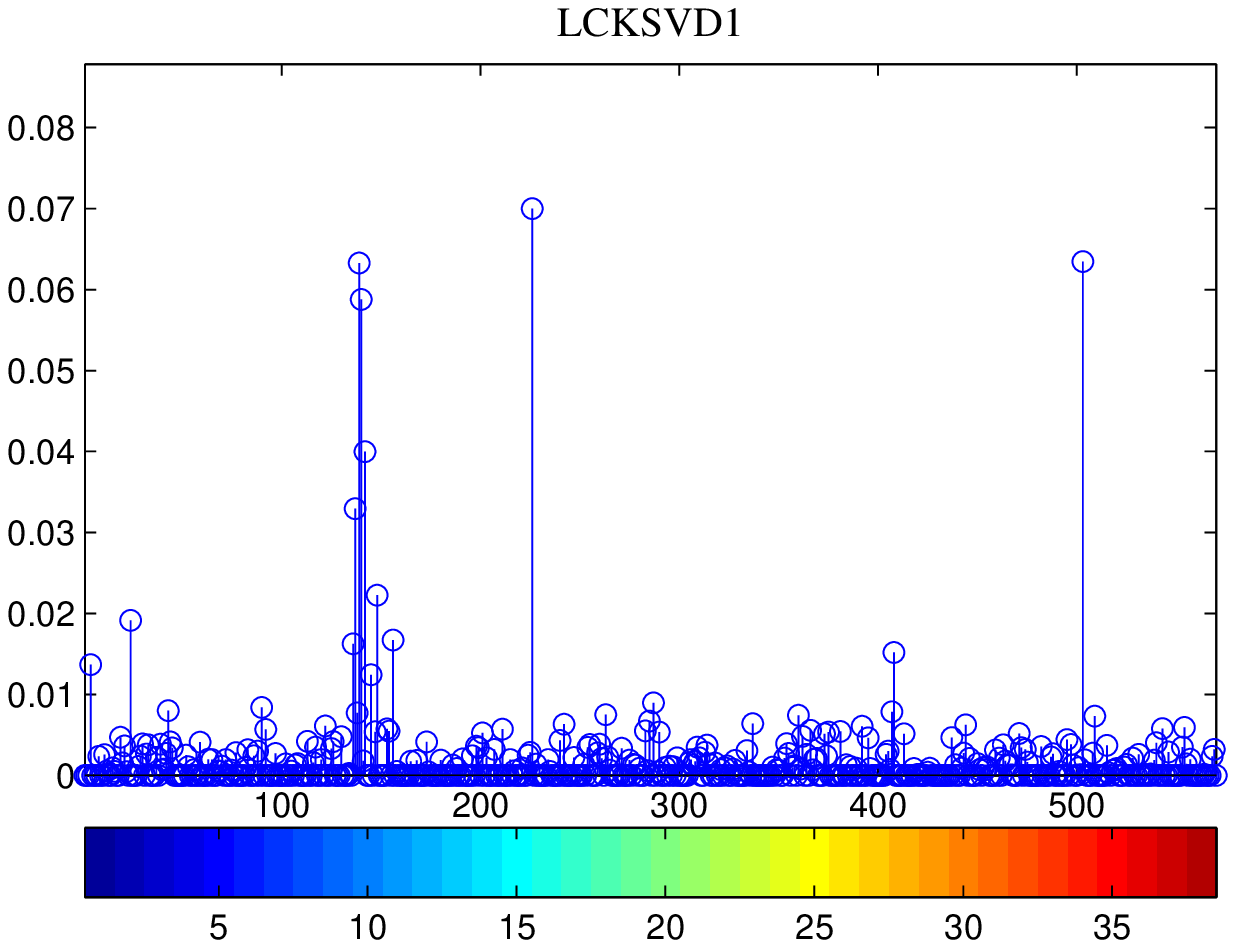}\label{fig:spcodes_LCKSVD1}} \hspace{-0.53cm}
\subfloat[]{\includegraphics[width=0.25\textwidth]{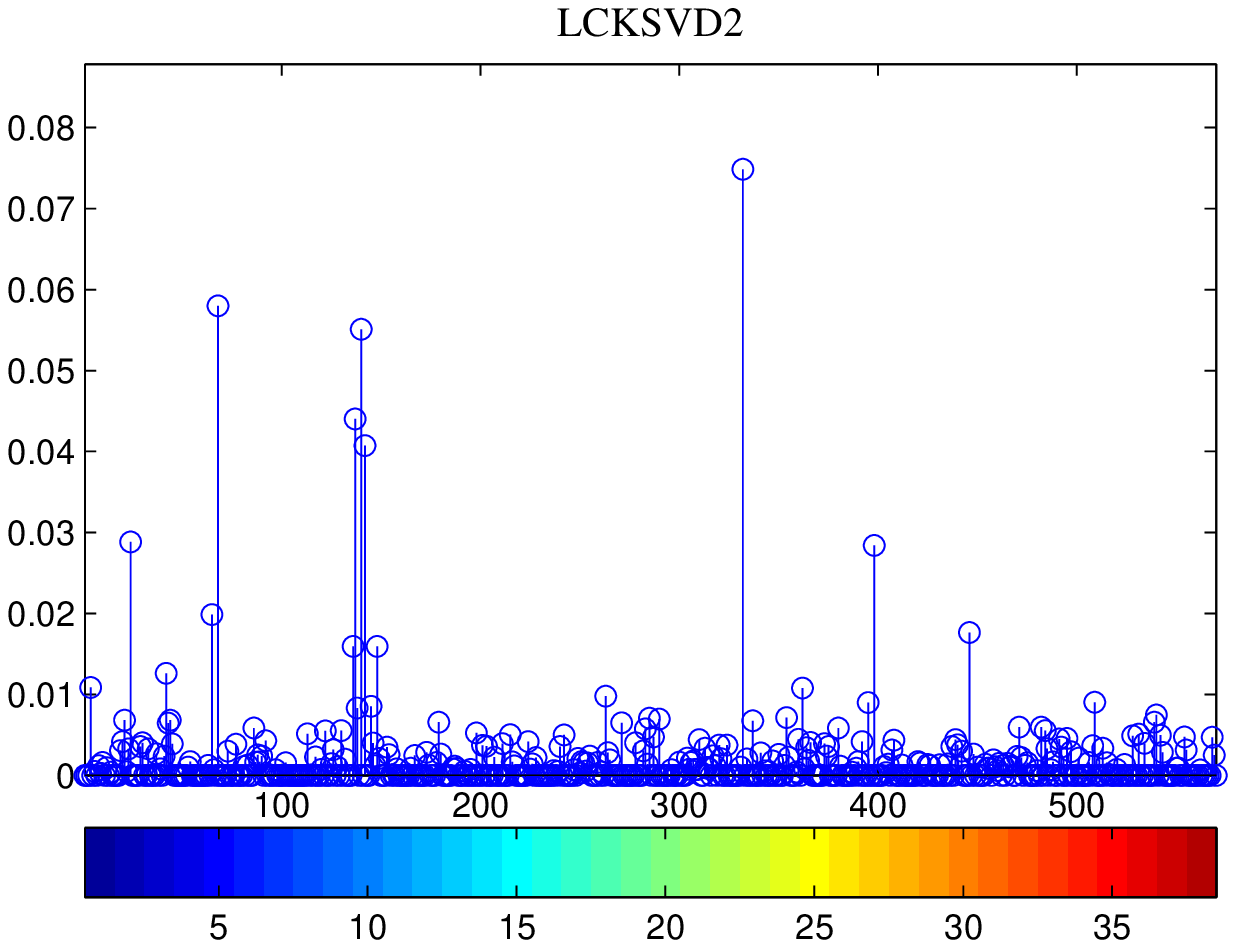}\label{fig:spcodes_LCKSVD2}} \hspace{-0.53cm}
\subfloat[]{\includegraphics[width=0.25\textwidth]{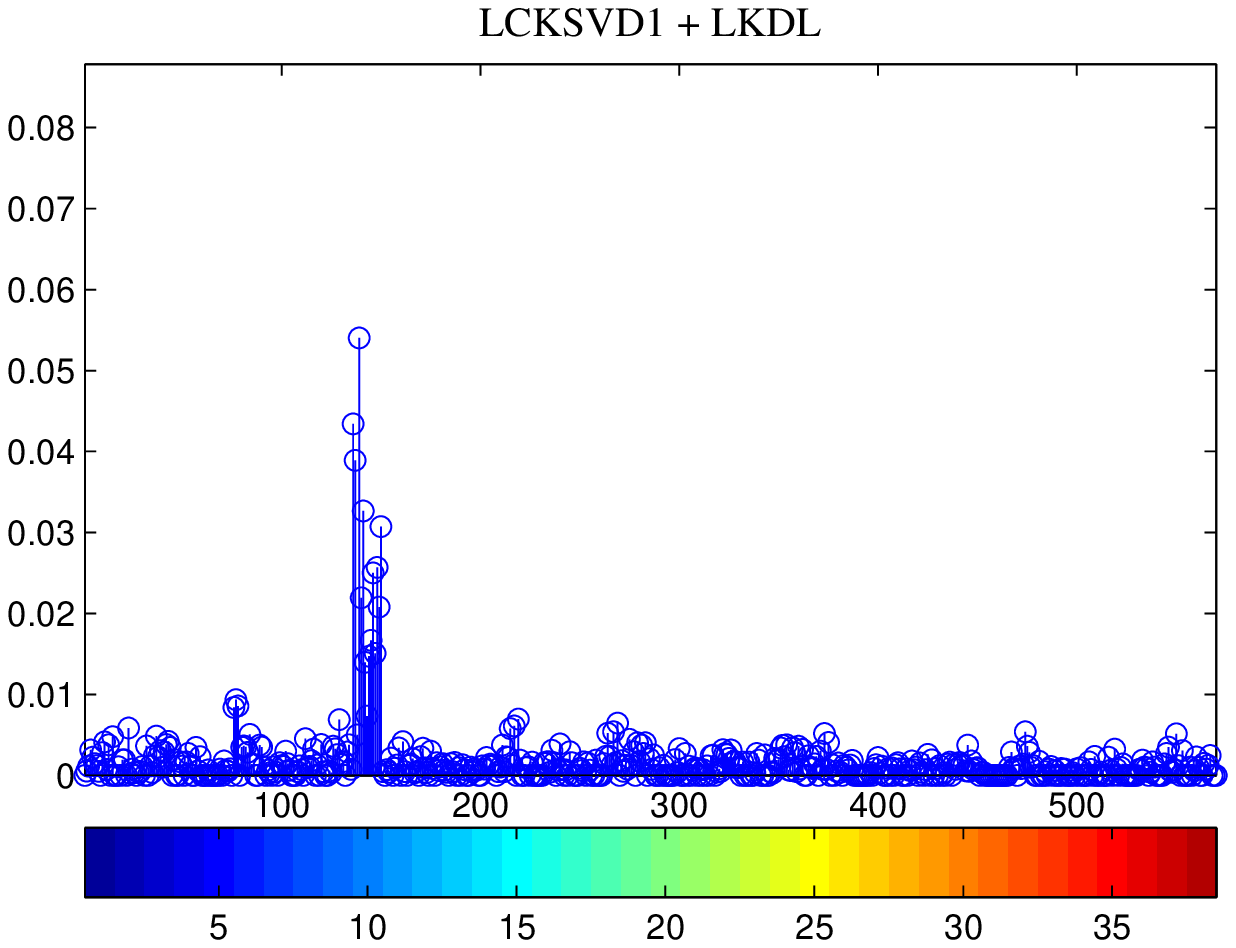}\label{fig:spcodes_LCKVD1_LKDL}} \hspace{-0.53cm}
\subfloat[]{\includegraphics[width=0.25\textwidth]{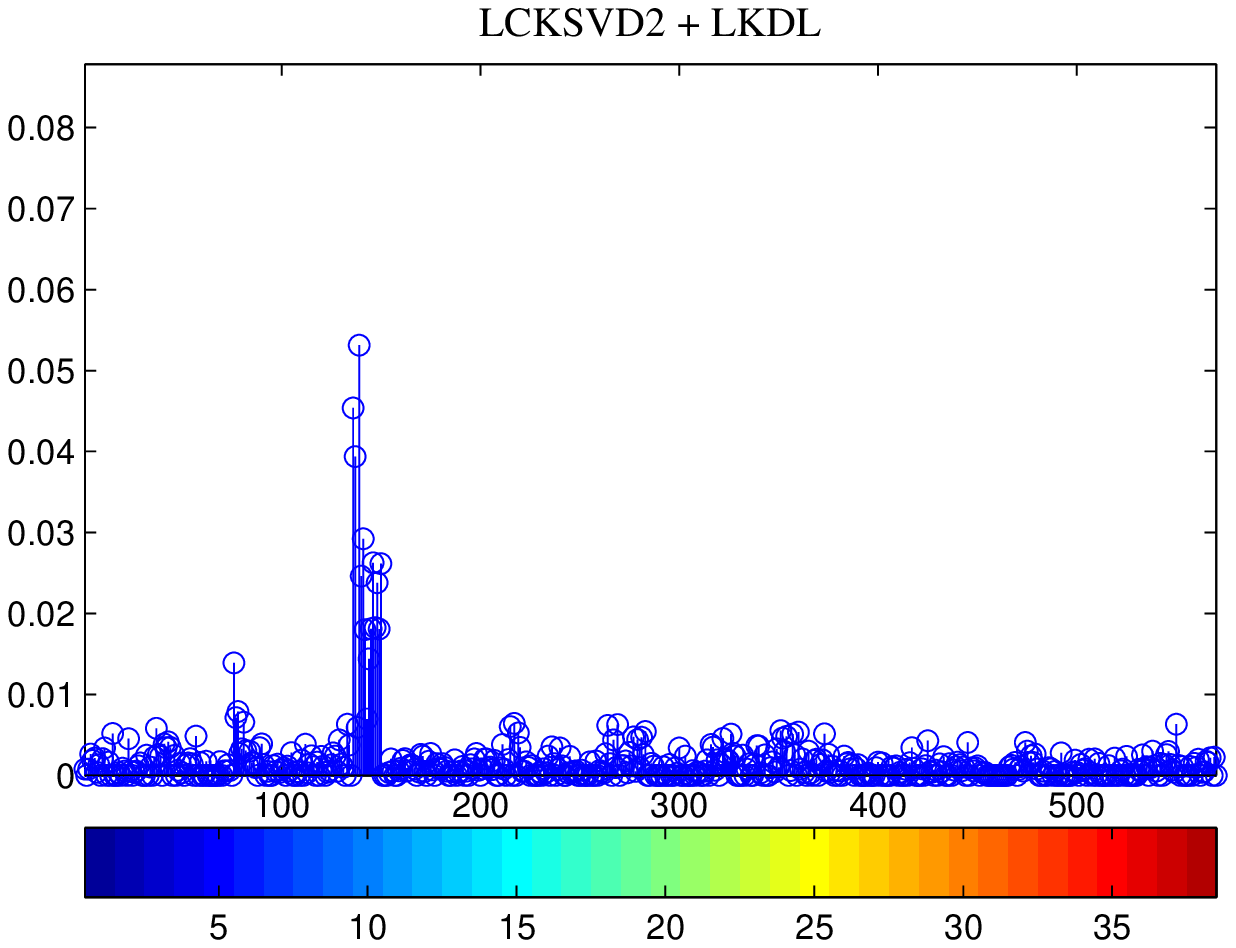}\label{fig:spcodes_LCKSVD2_LKDL}}
\vspace{-0.8cm}
\hspace{-0.5cm}
\subfloat[]{\includegraphics[width=0.25\textwidth]{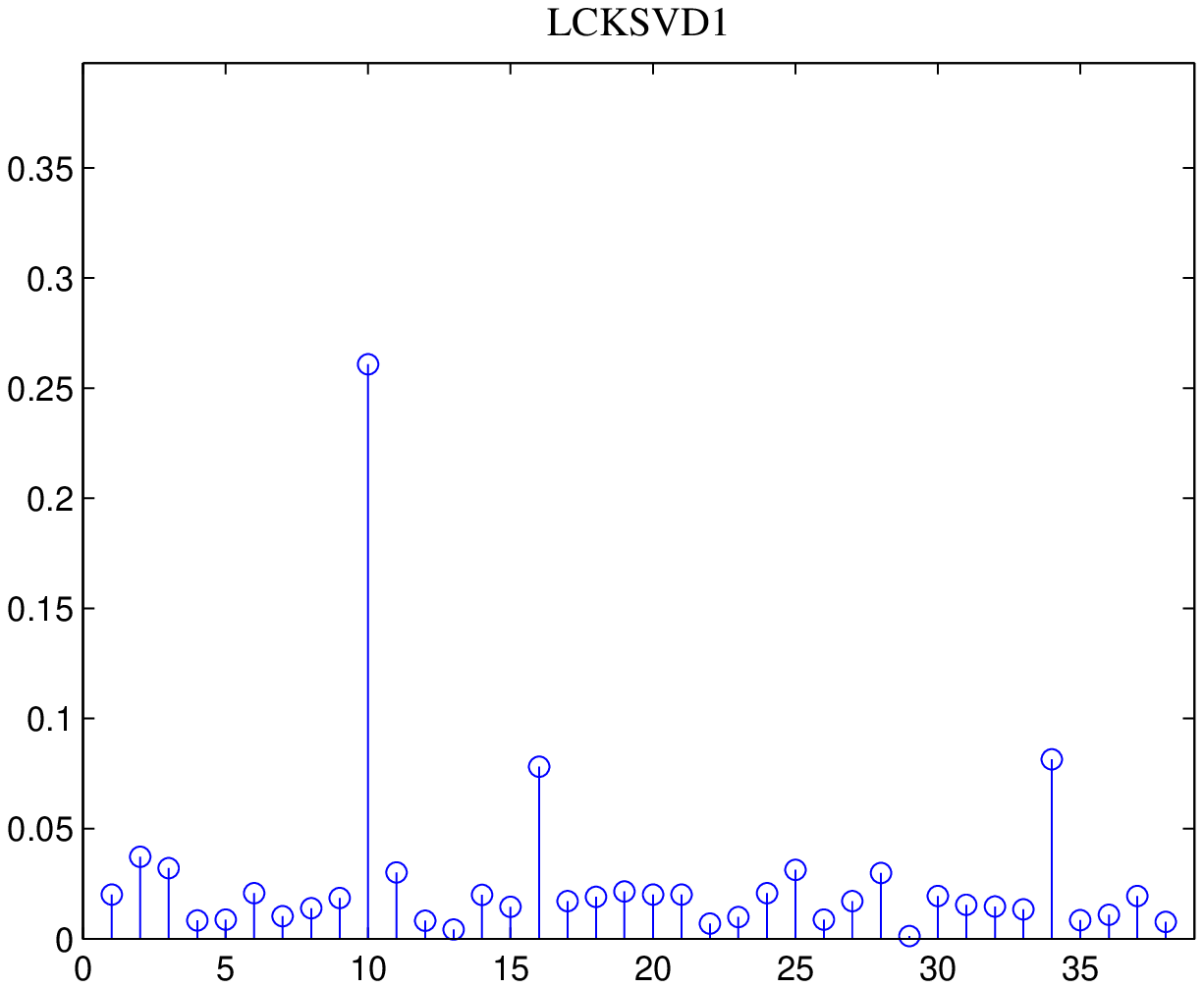}\label{fig:red_spcodes_LCKSVD1}} \hspace{-0.53cm}
\subfloat[]{\includegraphics[width=0.25\textwidth]{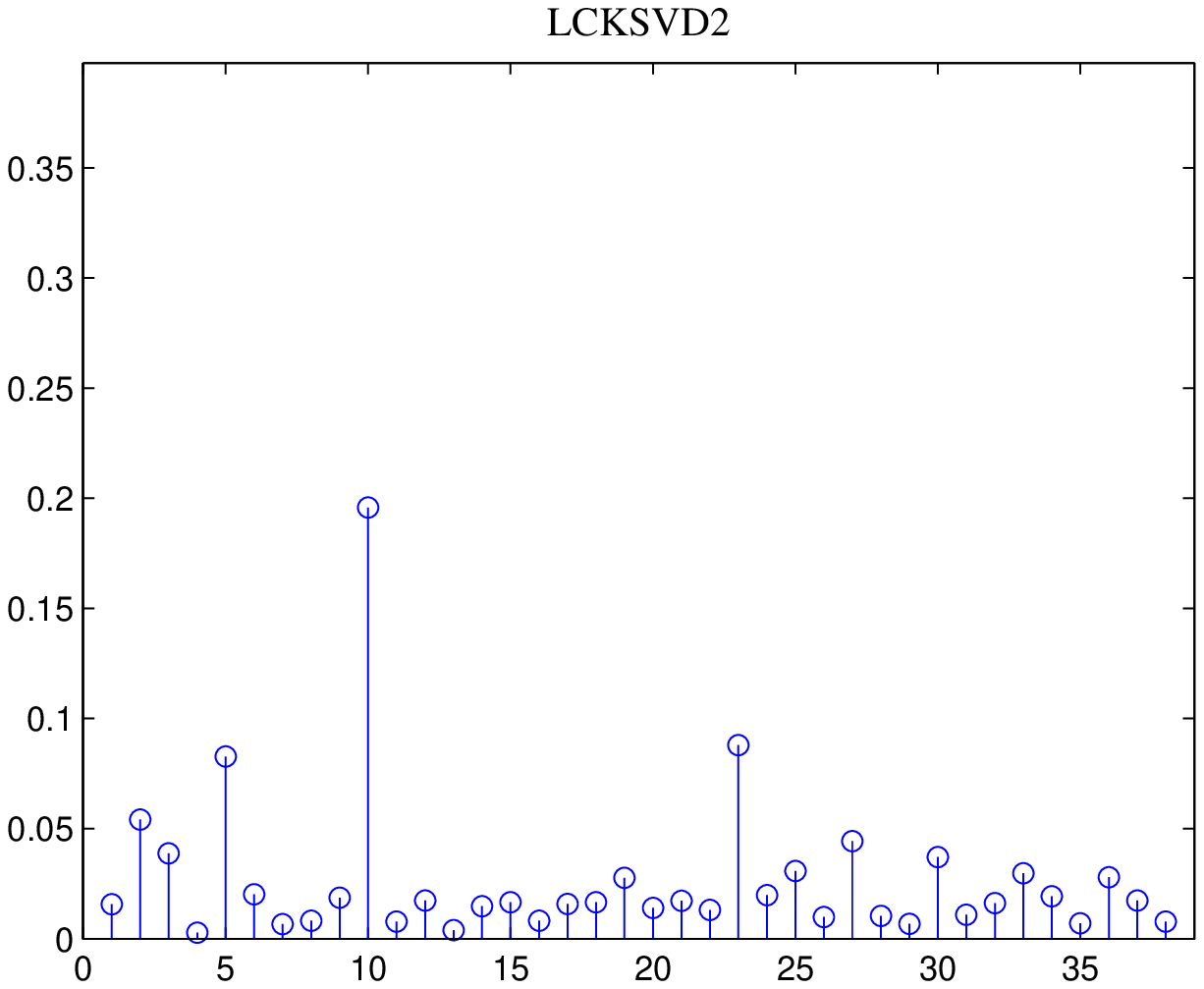}\label{fig:red_spcodes_LCKSVD2}} \hspace{-0.53cm}
\subfloat[]{\includegraphics[width=0.25\textwidth]{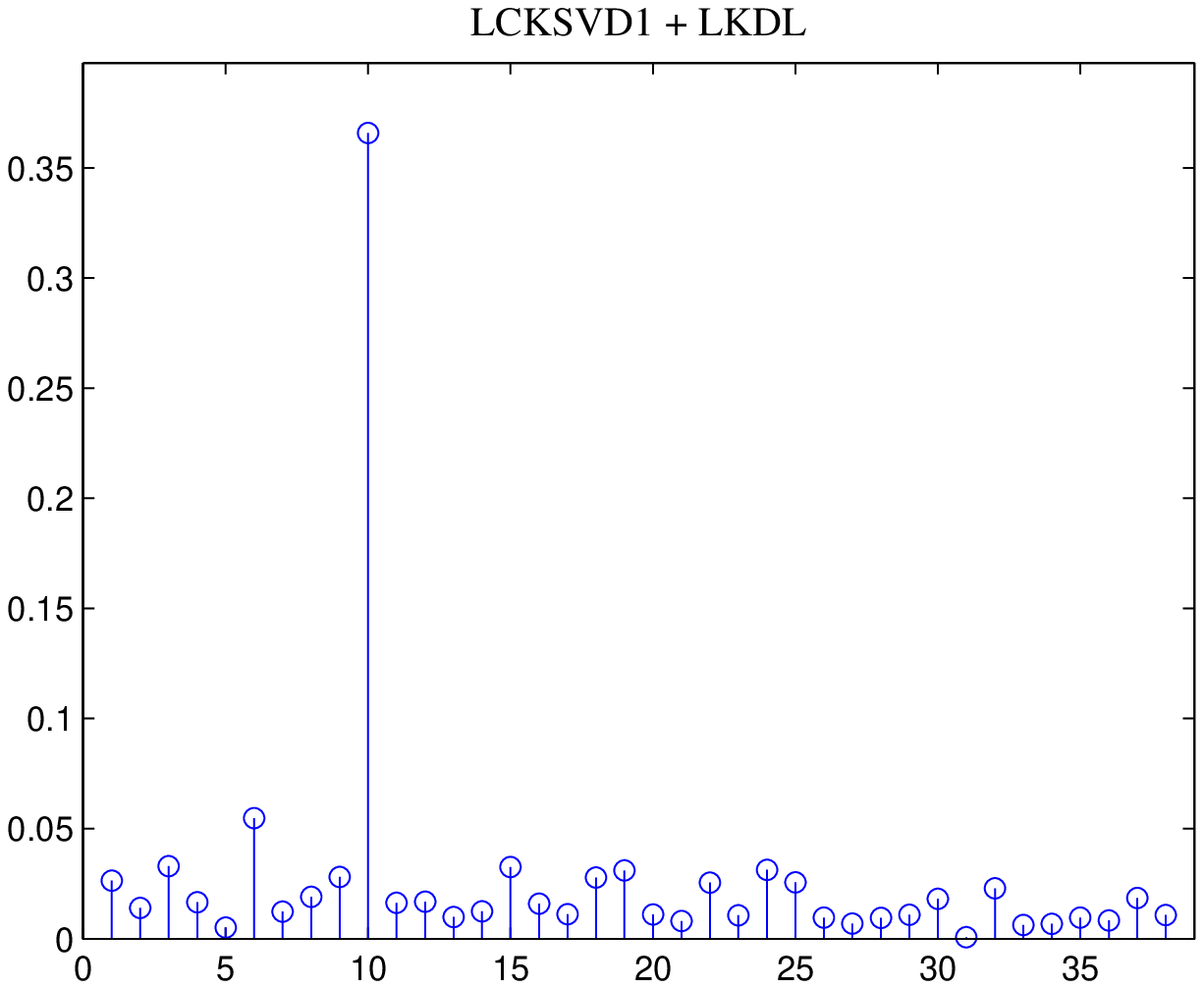}\label{fig:red_spcodes_LCKSVD1_LKDL}} \hspace{-0.53cm}
\subfloat[]{\includegraphics[width=0.25\textwidth]{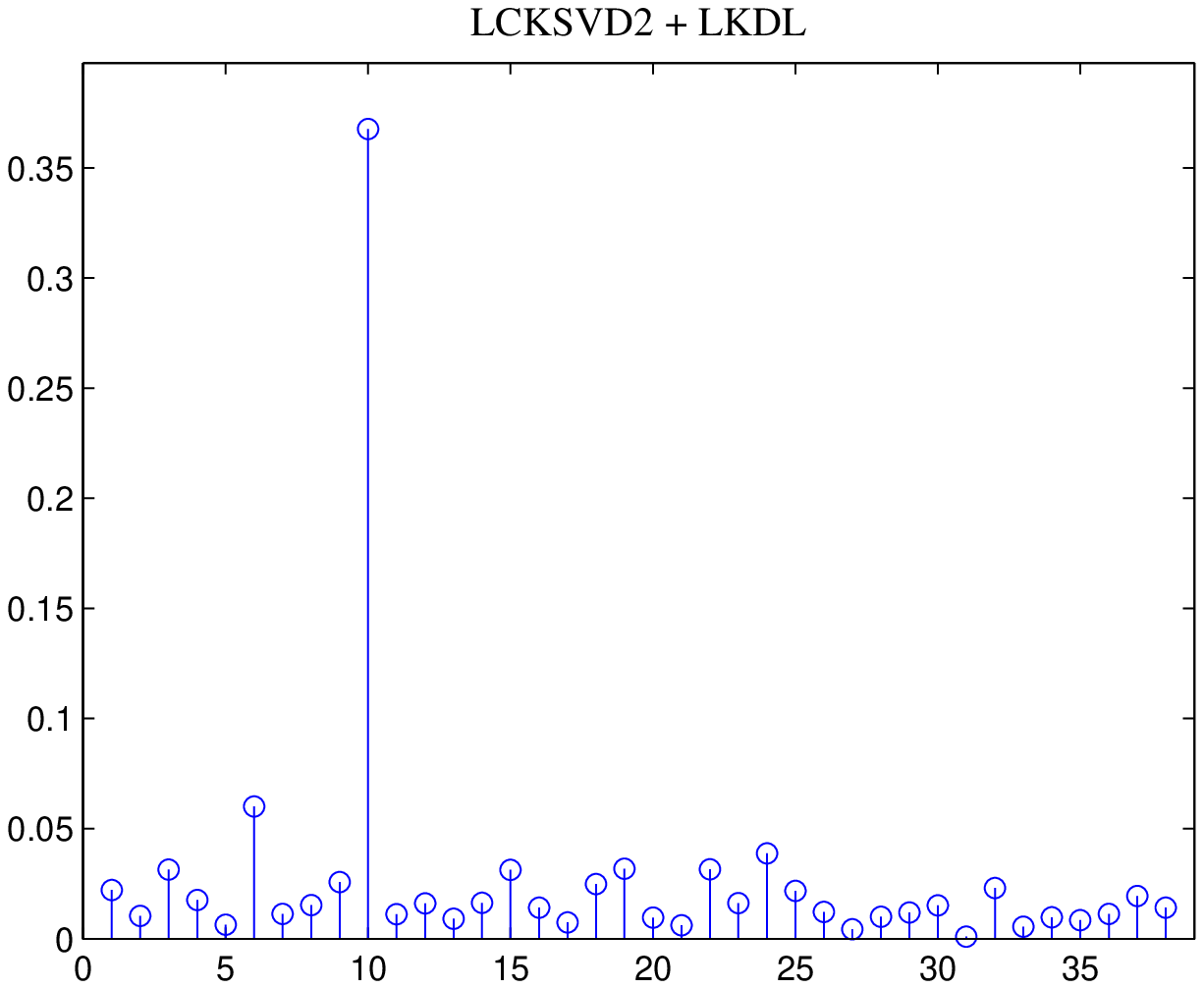}\label{fig:red_spcodes_LSKSVD2_LKDL}}
\vspace{-0.5cm}
\caption{Upper row: sum of absolute values of sparse coefficient vectors (of size 570, the size of the dictionary) corresponding to test examples from class `10' in Extended YaleB database. The columns from left to right represent LC-KSVD1 and LC-KSVD2 with and without the addition of LKDL pre-precessing. The additional colorbar features 38 bars which correspond to 38 classes in Extended YaleB. Bottom row: additional summation of the absolute values of sparse coefficients in every class. As expected, the majority of nonzero values in all sparse coefficient vectors originate from class `10'.}
\label{fig:GRAPH_4}
\end{figure*}

The improved discriminability of LKDL combined with LC-KSVD, versus LC-KSVD alone, can be demonstrated by inspecting the resulting sparse coefficients of the test set. In Fig. \ref{fig:GRAPH_4} one can see the obtained sum of absolute values of the sparse coefficient vectors of all 32 test samples from class `10'. The ideal distribution of atoms chosen during sparse-coding should be concentrated around atoms: $[139,\cdots,150]$, which belong to class `10'.
One can see that in the case of LC-KSVD, there are a few ``successful'' atoms which largely contribute to the reconstruction of the test samples, while in LC-KSVD combined with LKDL, the contribution is distributed more evenly between all of the atoms in that class. In addition, LC-KSVD alone will often choose atoms not corresponding with the given class, while in LKDL, the contribution of these atoms is fairly small.

\captionsetup[subfigure]{labelformat=empty}

\begin{figure*}[!t]
\centering
\subfloat[]{\includegraphics[width=0.45\textwidth]{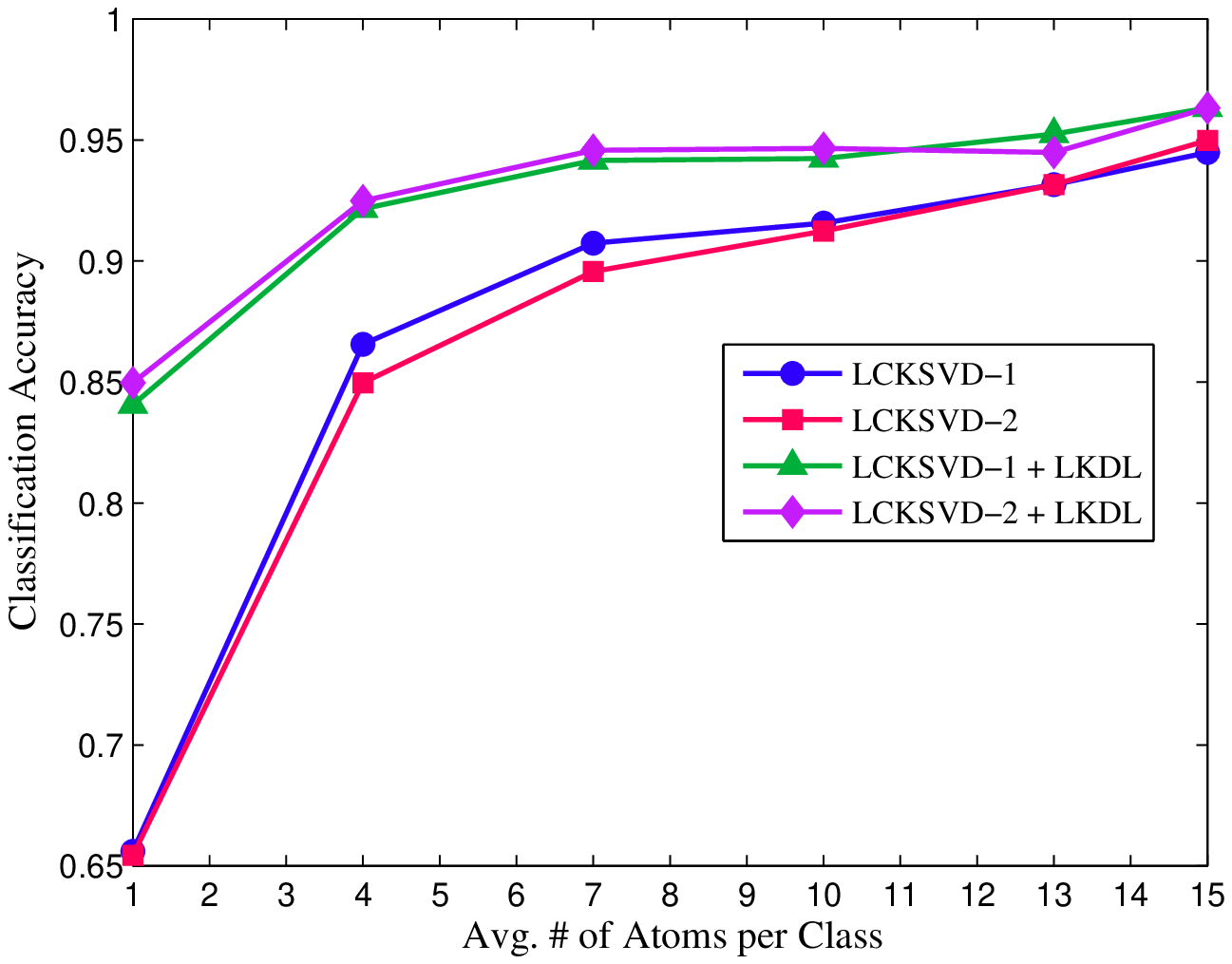}\label{fig:LCKSVD_numatoms}}
\hfil
\subfloat[]{\includegraphics[width=0.45\textwidth]{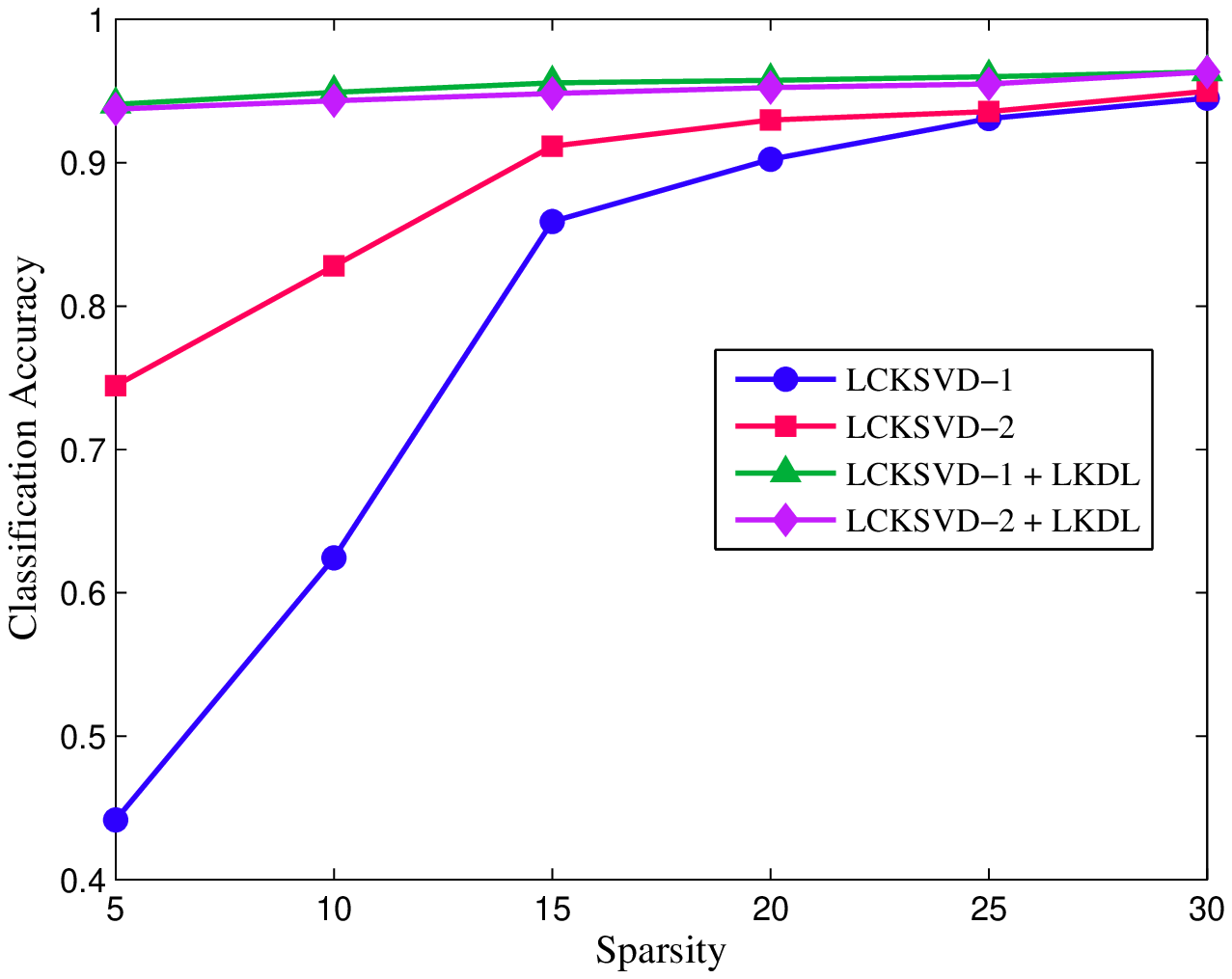}\label{fig:LCKSVD_sparsity}}
\caption{Dependance of accuracy in the average number of atoms per class (a) and the sparsity factor (b).}
\label{fig:GRAPH_5}
\end{figure*}

Next we explore the impact of LKDL on the size of the learned dictionary. Fig. \ref{fig:LCKSVD_numatoms} shows the results of LC-KSVD1 and LC-KSVD2, with and without LKDL, versus the average number of dictionary atoms for each class. It is clear that LKDL improves the results of both LC-KSVD1 and LC-KSVD2. With the addition of LKDL, a smaller dictionary with 7 atoms per person achieves the same results of LC-KSVD alone with 15 atoms per person. This gap in performance grows as the size of the dictionary becomes smaller and reaches a 20\% difference for 1 atom per person. The conclusion is that a more compact dictionary can be learnt using the combination of LC-KSVD and LKDL, without compromising accuracy.

Fig. \ref{fig:LCKSVD_sparsity} shows a similar experiment of the dependency of classification on the sparsity factor, i.e. the number of atoms used in the sparse reconstruction of a given signal. The combination of LKDL and LC-KSVD with a sparsity of 15 achieves a better accuracy than that of LC-KSVD alone with a sparsity of 30. From both these figures it can be seen that the addition of LKDL can be helpful in reducing the complexity of the DL problem, without compromising the accuracy.

\subsubsection{Evaluation on the AR Face Database}

The AR Face database consists of 4,000 color images of faces belonging to 126 classes. Each class consists of images taken over two sessions, containing different lighting conditions, facial variations and facial disguises (sunglasses and scarves). Following the experiment in \cite{LCKSVD2}, 2,600 images were chosen, first 50 classes of males and first 50 classes of females. Out of 26 images in each class, 20 were chosen for training and the rest for evaluation. We use the already-processed dataset\footnote{Found in \url{http://www.umiacs.umd.edu/~zhuolin/LCKSVD/}} in \cite{LCKSVD2}, where the original images of size $165 \times 120$ pixels were reduced to $540$-dimensional vectors using random projection as in Extended YaleB. The cardinality is same as before set to 30 and the number of atoms in DL is set to 500 (5 atoms per class). As before, we normalized all the signals to unit $l_2$-norm. The parameters $\sqrt{\alpha}$ and $\sqrt{\beta}$ were determined using coarse-to-fine 5-fold cross validation. We have noticed that an optimal parameter of $\sqrt{\alpha}$ for LC-KSVD1 is not necessarily as good for LC-KSVD2, thus we chose two sets of parameters: $\sqrt{\alpha}=\sqrt{\beta}=1/150$ for the optimal result of LC-KSVD1 (the value of $\beta$ is not really used in LC-KSVD1), and $\sqrt{\alpha}=\sqrt{\beta}=1/120$ for LC-KSVD2.

In table \ref{table:LCKSVD AR_performance} we compare the classification results of LC-KSVD1 and LC-KSVD2, with and without LKDL pre-processing. As can be seen our method improves the performance of LC-KSVD1 by 1.5\% and LC-KSVD2 by 1\%.

\begin{table}[!t]
\caption{Classification accuracy of LC-KSVD1 and LC-KSVD2 on the AR Face database, with and without LKDL pre-processing}
\label{table:LCKSVD AR_performance}
\centering
\begin{tabular}{||l||c||}
\multicolumn{1}{l}{\bf Algorithm}  &\multicolumn{1}{c}{\bf Accuracy} \\
\hline
LC-KSVD1 & 92.5 \\
\hline
LC-KSVD1 + LKDL  & \textbf{94} \\
\hline
\hline
LC-KSVD2 & 93.7 \\
\hline
LC-KSVD2 + LKDL  & \textbf{94.7} \\
\hline
\end{tabular}
\end{table}

\section{Conclusion} \label{Conclusion}

In this paper we have discussed some of the problems arising when trying to incorporate kernels in DL, and payed special attention to the kernel-KSVD algorithm by Nguyen \textit{et al.} \cite{KDL,KDL2}.
We proposed a novel kernel DL scheme, called ``LKDL'', which acts as a kernelizing pre-processing stage, before performing standard DL.
We used the concept of virtual training and test sets and described the different aspects of calculating these signals.
We demonstrated in several experiments on different datasets the benefits of combining our LKDL pre-processing stage, both in accuracy of classification and in runtime.
Lastly, we have shown the easiness of integrating our method with existing supervised and unsupervised DL algorithms. It is our hope that the proposed methodology will encourage users to consider kernel DL for their tasks, knowing that the extra-effort involved in incorporating the kernel layer is near-trivial. We intend to freely share the code that reproduces all the results shown in this paper.

Our future research directions include combining LKDL with online DL.
We would also like to examine the benefit of applying LKDL to the sparse coefficients instead of the input signals and maybe combining both options.
Lastly, our goal is improving the sampling ratio, i.e. the size of the matrix $\bC$, using more advanced sampling techniques.



%



\ifCLASSOPTIONcaptionsoff
  \newpage
\fi



\bibliographystyle{IEEEtran}
\bibliography{IEEEabrv,final_bib}

%

\vspace*{-30pt}

\begin{IEEEbiography}[{\includegraphics[width=1in,height=1.25in,clip,keepaspectratio]{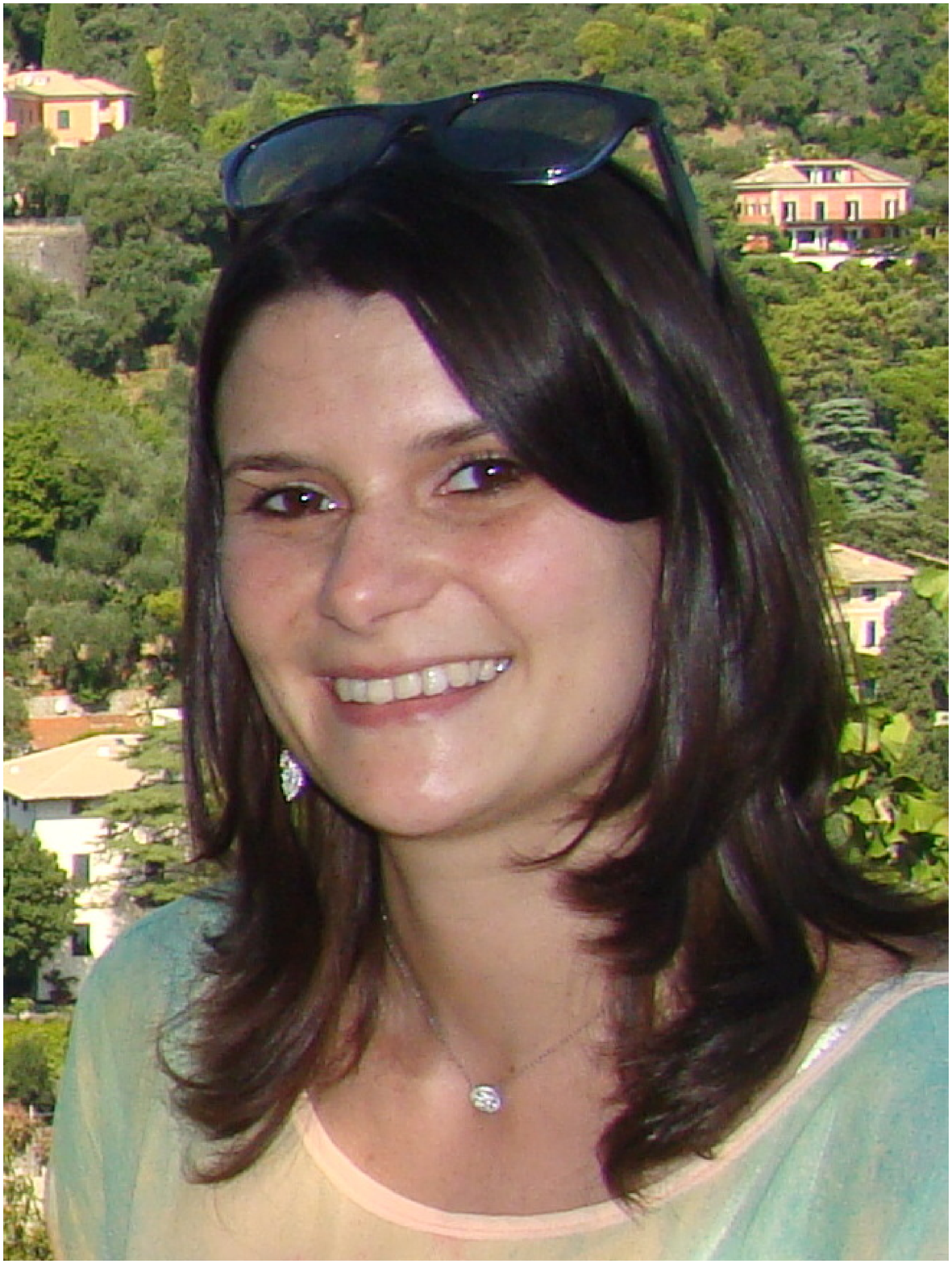}}]{Alona Golts}
received her B.Sc. (2009) in Electrical Engineering and Physics,
from the department of Electrical Engineering at the Technion,
Israel, where she is currently pursuing her M.Sc degree.
Alona has served in the Israeli Air Force from 2009 to 2015, under the reserve excellence program ``Psagot''.
\end{IEEEbiography}

\vspace*{-30pt}

\begin{IEEEbiography}[{\includegraphics[width=1in,height=1.25in,clip,keepaspectratio]{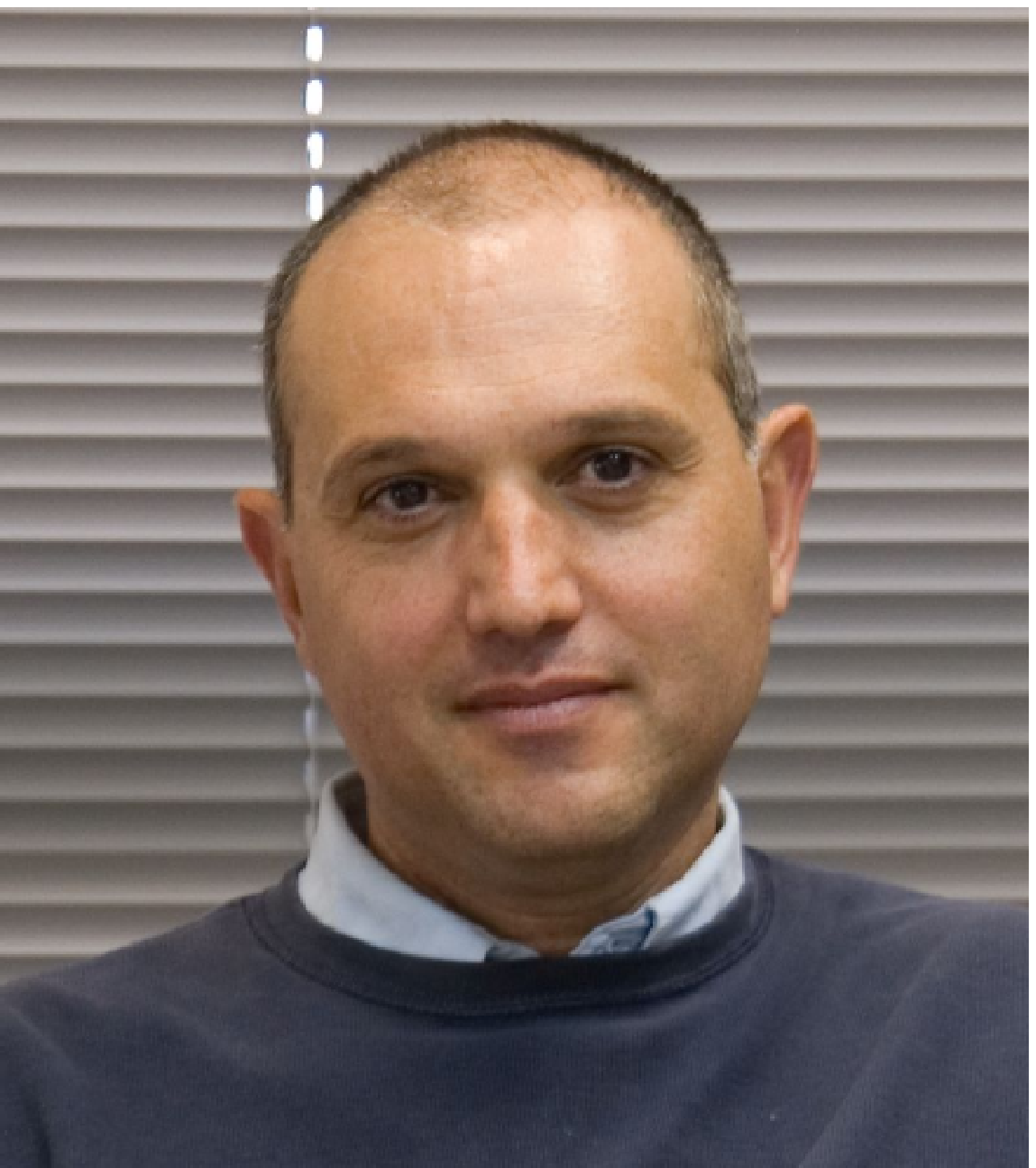}}]{Michael Elad}
received his B.Sc. (1986), M.Sc.
(1988) and D.Sc. (1997) from the department of
Electrical engineering at the Technion, Israel. Since
2003 he is a faculty member at the ComputerScience
department at the Technion, and since 2010
he holds a full-professorship position.
Michael Elad works in the field of signal and
image processing, specializing in particular on inverse
problems, sparse representations and superresolution.
Michael received the Technion’s best
lecturer award six times, he is the recipient of the
2007 Solomon Simon Mani award for excellence in teaching, the 2008 Henri
Taub Prize for academic excellence, and the 2010 Hershel-Rich prize for
innovation. Michael is an IEEE Fellow since 2012. He is serving as an
associate editor for SIAM SIIMS, and ACHA. Michael is also serving as
a senior editor for IEEE SPL.
\end{IEEEbiography}

\vfill
\vfill
\vfill
\vfill
\vfill
\vfill
\vfill
\vfill





\end{document}